Building and Testing a General Intelligence Embodied in a Humanoid Robot
Version 1.0, 2023-07-26

Suzanne Gildert and Geordie Rose
Sanctuary Cognitive Systems Corporation

July 26, 2023

## Abstract

Machines with human-level intelligence should be able to do most economically valuable work. This aligns a major economic incentive with the scientific grand challenge of building a human-like mind. Here we describe our approach to building and testing such a system. Our approach comprises a physical humanoid robotic system; a software based control system for robots of this type; a performance metric, which we call g+, designed to be a measure of human-like intelligence in humanoid robots; and an evolutionary algorithm for incrementally increasing scores on this performance metric. We introduce and describe the current status of each of these. We report on current and historical measurements of the g+ metric on the systems described here.

## 1. Introduction

How are we to understand how our own minds work? We take the position that the process of building a mind like ours is necessary to gain this understanding. Nils J. Nillson wrote "I'm not as interested in manufacturing intelligence as I am in understanding it. But I think we'll need to manufacture it before we can understand it." (Nilsson, 2005). We adopt this stance here.

What is intelligence, and how is it related to the broader question of how minds work? We use McCarthy's definition, where intelligence is "the computational part of the ability to achieve goals in the world" (McCarthy, 2007). This definition makes intelligence instrumental to the more profound process of changing the world to the benefit of the mind possessing that intelligence.

Why should a mind seek certain goals over others? For biological minds, goals are at least in part naturally selected by evolution. For engineered minds, we get to choose what they are. One way to do this is by using the reward paradigm in reinforcement learning (Sutton, 2018).

There are many different sorts of goals an engineered mind could seek. Perhaps at some level of abstraction all minds are variants of the same basic architecture, regardless of what goals an entity may favor. Searching for such an architecture is one of the goals of the reinforcement learning (Silver, 2021) and artificial general intelligence programs (Legg, 2007).

Here we will limit our scope to focus specifically on human-like minds. What problems does the world present to us? Instead of confronting this directly, we ask a related question. Paraphrasing (McCarthy, 1996): What problems would the world present to a humanoid robot faced with the problems we might be inclined to relegate to sufficiently intelligent robots of this kind?

Answering this much narrower question is still difficult, but here our intuitions may be more clear. One of the obvious uses of such robots, should they be possible to make, is to do work. From (Nilsson, 2005):

> *I claim that achieving real human-level artificial intelligence would necessarily imply that most of the tasks that humans perform for pay could be automated… Machines exhibiting true human-level intelligence should be able to do many of the things humans are able to do. Among these activities are the tasks or "jobs" at which people are employed. I suggest we replace the Turing test by something I will call the "employment test." To pass the employment test, AI programs must be able to perform the jobs ordinarily performed by humans. Progress toward human-level AI could then be measured by the fraction of these jobs that can be acceptably performed by machines.*

The idea that progress towards human-level AI could be quantified by an "employment test" is powerful for several reasons:

1. It aligns a major economic incentive (automating labor is by definition the largest market possible) with the scientific grand challenge of understanding the human mind. Lack of alignment between economic value and understanding minds, which includes but isn't limited to general intelligence, has been an issue in the field of AI tracing back to its early days. There are many incentives for researchers, technologists, and investors to focus on what McCarthy called "bounded information theoretic situations" (McCarthy, 2007) — special purpose solutions to specific problems. That it may be possible, even in principle, to build a new technological category that can do most work — call it *autonomous labor*, although what is meant by that is work in general, not just the sorts of tasks conventionally associated with automation — makes the calculus of investment in such a project an interesting exercise, which largely boils down to predicting when human-level AI will be achieved.

2. It provides the potential for a continuous gradient of progress, where cumulative integration of small improvements can build over time to success. Using performance of work as a frame, we can create measurement systems that allow initially non-performant systems to slowly and gradually get better over time, and in the best case, make progress towards human-level AI predictable. It is a truism that "what's measured improves". Having a way to quantify progress towards a long-term grand challenge objective is critically important to eventual success, helping to avoid (paraphrasing LeCun) off-ramps on the highway towards human-level AI (LeCun, 2023).

3. Direct measurement of intelligence in machines (and people for that matter) is fraught with difficulties. Being able to measure whether or not a machine can do work is not completely straightforward, but it is arguably possible to do so in a way that is good enough to establish a gradient of progress towards human-level AI.

4. There is a rich history of support for the notion that a human-like mind is best situated as the control system of a humanoid robot (Pfeifer, 2006). In the "employment test" frame, the benefit of a human-like embodiment is incontrovertible. The world of people is built around objects, spaces, and events that are specifically engineered for the human form. While of course it is possible to build useful machines that are not human-like, even ones that could have powerful minds and capabilities, they would not perform as well as machines with human-like forms on a suitably general portfolio of work.

Given the above framing of the problem, we are still left with many decisions about what specifically to build. Our approach is similar in spirit to building what Nilsson called *habile systems* (Nilsson, 1995).

The philosophy of this approach is to separate the problem into two parts:

1. The design and construction of a core entity inside which prior (unlearned) information resides. This core is somewhat analogous to the innate capabilities we are born with deriving from our evolutionary heritage. This includes a choice of physical embodiment (in our case, the robot, including its sensors, actuators, onboard computers, communications networks, and the like), and a specific cognitive architecture acting as an autonomous control system for that robot. The idea of starting with a core entity, which has been called a 'child machine', traces back to Turing and the foundations of the AI field (Turing, 1950). What might be good ideas to include as priors has been discussed at length (Turing, 1950, McCarthy, 1999). Note that 'child machine' does not mean that the core system is literally like a human child. It just means a system imbued with a set of priors that allows the second step below to proceed. Computational systems have different natural strengths and weaknesses in relation to human minds; when possible we would like to imbue machines with superhuman capabilities related to achieving general work goals (for example, arithmetic, endurance and memory). Sutton (Sutton, 2019) has argued that priors of this kind have historically not been helpful, and we should strive to build systems where step 1 is minimized and step 2 below takes center stage; however some priors (such as modes of sensing and acting) are required. There is only so far that the bitter lesson can be pushed before choices must be made by (hopefully intelligent) designers.

2. An exploration and learning process where the core entity, with whatever priors we have chosen to imbue it with, is exposed to the world and learns to perform increasingly complex tasks.

A human-like mind must have its own independent agency of the sort we ourselves have. Using Searle's terminology, we seek to build a strong AI (Searle, 1980):

> *According to weak AI, the principal value of the computer in the study of the mind is that it gives us a very powerful tool… But according to strong AI, the computer is not merely a tool in the study of the mind; rather, the appropriately programmed computer really is a mind.*

Weak AI systems (those designed to be tools for use by people) can of course also be used by non-human strong AI systems, should they be possible to build.

Searle himself was a vocal advocate for the weak AI position, and against the strong AI position. He went so far as to develop an argument that strong AI was impossible (with the overtone that we shouldn't want to achieve it anyway), and embraced weak AI as appropriate, doable, and socially acceptable. We find echoes of these arguments in 'AI alignment' arguments (Yudkowsky, 2016). Whether Searle was right about strong AI being impossible is ultimately an empirical question. While there are good reasons to dismiss his arguments based on ourselves being both machines and minds and the fungibility of information (Landauer, 1999), it has been argued that this may not be enough (Brooks, 2021).

The remainder of this paper is structured as follows. In Section 2, we describe a methodology for defining and measuring the intelligence of a humanoid robot. In Section 3 we describe the type of humanoid robot we focus on here. In Section 4 we describe the autonomous control system we used, including both the underlying architecture of the core system and a novel evolutionary learning paradigm. In Section 5 we define and measure task performance of this system, and use the scoring system defined in Section 2 to extract the figures of merit defined there. We conclude in Section 6 by discussing limitations of the approach presented here.

## 2. An Employment Test for Robots

In 2019 we began developing an approach to measuring the capability of a robot to do work derived from the O*NET (Occupational Information Network) system (Gildert, 2019). O*NET is a source of information about work and workers developed and maintained by the U.S. Department of Labor/Employment and Training Administration (USDOL/ETA) (ONET, 2023).

This approach is predicated on the notion that economic activity forms a hierarchy of goals that overlaps significantly with the set of all human goals. In our framing of the problem of building a human-like mind, we make the non-trivial assumption that a machine that can do anything we value enough to pay for must possess a mind with many, perhaps all, of the properties of our own minds. For example, Marcus provides a threshold for a system to be considered AGI, which would easily be passed by a system that could do all work (Marcus, 2022).

What follows does not require agreement with the premise that having a human-like mind and ability to do work are closely related. But it was the motivation for developing this approach in the first place. While it is of immense practical importance to build a general purpose technology that can do most work, the real reason we are interested in this is understanding what it takes to design and incrementally improve a mind capable of performing that work.

### 2.1. O*NET's Characterization of Work

O*NET has two separate but related parts. The first quantifies *work*; what occupations there are in the US economy, and what are their requirements and definitions. The second quantifies the *worker*; what attributes are required for people to be able to participate in the economy. We begin by reviewing how O*NET quantifies work.

There are approximately one thousand distinct *occupations* identified by O*NET. An occupation is a job type, such as Electrician, Retail Salesperson, or Computer Hardware Engineer. O*NET purports to contain within it most of the occupations in the US economy (ONET, 2023a).

Each occupation is described with reference to a set of *tasks.* Tasks are the most detailed and concrete descriptions of the subcomponents of occupations available in O*NET. Currently there are 19,265 identified tasks (ONET, 2023b). Each task in O*NET is labeled by an integer, ranging from 1 ("Resolve customer complaints regarding sales and service") to 23,955 ("Dismount, mount, and repair or replace tires"). Some integers in this range are not associated with tasks. Tasks arise from surveys asking practitioners of each occupation what activities must be performed to be successful at the occupation. In addition to text descriptions, each task is scored as to its importance with a maximum value of 100, and whether it is core (required) or supplemental (nice to have).

Occupations typically require the performance of 20-30 different tasks in order to do everything the job requires. Selecting for example the Retail Salespersons occupation, we obtain 24 tasks (see Table 1) (ONET, 2023c). O*NET contains similar breakdowns for every occupation tracked.

O*NET also includes a taxonomy where tasks are mapped into more abstract categories called *work activities* (ONET, 2023d). Unlike tasks, which are specifically written by practitioners of specific occupations for those occupations, work activities were designed by USDOL/ETA employees to capture occupation-independent similarities between reported tasks.

Work activities come in five hierarchical levels of abstraction — detailed, intermediate, and three general levels. Each detailed work activity, of which there are 2,068, is part of a broader class (334 intermediate), and the intermediate ones are parts of yet broader classes (41 general). These 41 general work activities feed into nine and finally four at the topmost level. Labeling for work activities follows the O*NET hierarchical content model labeling scheme [(ONET, 2023e)](); see also Figure 1.

| Importance | Category | Task |
|---|---|---|
| 96 | Core | Greet customers and ascertain what each customer wants or needs. |
| 92 | Core | Recommend, select, and help locate or obtain merchandise based on customer needs and desires. |
| 89 | Core | Compute sales prices, total purchases, and receive and process cash or credit payment. |
| 89 | Core | Prepare merchandise for purchase or rental. |
| 88 | Core | Answer questions regarding the store and its merchandise. |
| 87 | Core | Maintain knowledge of current sales and promotions, policies regarding payment and exchanges, and security practices. |
| 83 | Core | Demonstrate use or operation of merchandise. |
| 83 | Core | Describe merchandise and explain use, operation, and care of merchandise to customers. |
| 82 | Core | Ticket, arrange, and display merchandise to promote sales. |
| 81 | Core | Inventory stock and requisition new stock. |
| 81 | Core | Exchange merchandise for customers and accept returns. |
| 77 | Core | Watch for and recognize security risks and thefts and know how to prevent or handle these situations. |
| 69 | Core | Place special orders or call other stores to find desired items. |
| 68 | Core | Clean shelves, counters, and tables. |
| 90 | Supplemental | Maintain records related to sales. |
| 86 | Supplemental | Open and close cash registers, performing tasks such as counting money, separating charge slips, coupons, and vouchers, balancing cash drawers, and making deposits. |
| 84 | Supplemental | Prepare sales slips or sales contracts. |
| 80 | Supplemental | Estimate and quote trade-in allowances. |
| 79 | Supplemental | Bag or package purchases and wrap gifts. |
| 79 | Supplemental | Help customers try on or fit merchandise. |
| 74 | Supplemental | Sell or arrange for delivery, insurance, financing, or service contracts for merchandise. |
| 69 | Supplemental | Estimate quantity and cost of merchandise required, such as paint or floor covering. |
| 57 | Supplemental | Rent merchandise to customers. |
| 56 | Supplemental | Estimate cost of repair or alteration of merchandise. |

**Table 1.** Fourteen core and ten supplemental O*NET tasks required for occupation 41-2031.00 - Retail Salespersons [(ONET, 2023c)]().

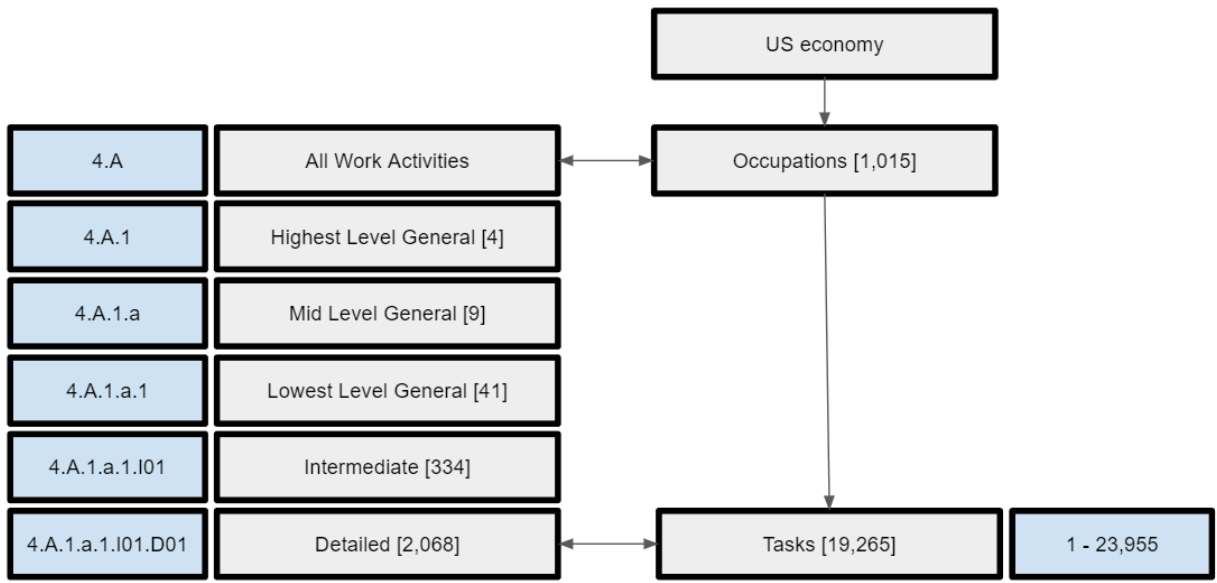

**Figure 1.** Work Activity, Task and Occupation hierarchies, with numbers of each in O*NET, and indicative labeling conventions (in blue). O*NET provides maps between Detailed Work Activities and Tasks (ONET, 2023f).

As an example of how the work activity hierarchy is labeled and structured, the detailed work activity "Gather customer or product information to determine customer needs", is labeled in the O*NET convention as 4.A.1.a.1.I14.D02. This means it is an example (D02, D standing for 'detailed') of the intermediate work activity with label 4.A.1.a.1.I14 ("Collect data about consumer needs or opinions", I14, I standing for 'intermediate') which in turn is part of the general work activity 4.A.1.a.1 ("Getting Information"). The remaining part of the label indicates inclusion in a more general work activity for each element of the label (4.A.1.a is "Looking for and Receiving Job-Related Information", 4.A.1 is "Information Input"). The top four levels of the work activity hierarchy are labeled 4.A.1 - 4.A.4; see (ONET, 2023d).

Tasks are descriptions of work required to perform specific occupations, whereas detailed work activities are occupation-independent generalizations of tasks. O*NET provides maps between tasks and detailed work activities (ONET, 2023f). For example, the most important task in the Retail Salesperson task list, "Greet customers and ascertain what each customer wants or needs", maps to two different detailed work activities, namely "Gather customer or product information to determine customer needs" and "Greet customers, patrons, or visitors".

## 2.2. O*NET's Characterization of Workers

In addition to characterizing work, O*NET also characterizes the properties of individual workers. This is done using a set of *work primitives*, which come in three variants. The first are *skills* (ONET, 2023h, ONET, 2023i). Skills are learned patterns of conditional actions; examples include mathematics, negotiation, and tending machines. The second are *abilities* (ONET, 2023g). These are innate properties, and include cognitive, physical, psychomotor and sensory abilities. The third type is *knowledge* (ONET, 2023j). Knowledge is also learned, and is defined as organized sets of principles and facts applying in general domains, including as examples arts and humanities, education, and law.

In the O*NET taxonomy, there are 33 skill types, 52 abilities, and 35 categories of knowledge; all of these are scored on a scale from 0 (none) to 7 (near the top of human performance). Lists of O*NET skills, abilities, and knowledge are included in the *O*NET Skills*, *O*NET Abilities*, and *O*NET Knowledge* tabs in [(Data, 2023)](#).

In the O*NET paradigm, these 33+52+35=120 numbers are sufficient to describe all of the underlying characteristics that a person can have relevant to work. We call the vector of these 120 numbers, representing a person's capabilities, their *work fingerprint*. A person's work fingerprint characterizes and individuates them fully in terms of what they can and can't do in the context of work.

The work fingerprint is O*NET's machinery for inferring whether a person has the *capability* of performing tasks, not whether any particular task *has actually been done* by that person. For example, from your own work fingerprint, one might be able to infer that you could do the work of a Retail Salesperson, but perhaps not a Dentist or an Anaesthesiologist, regardless of whether or not you'd actually tried to do those jobs. The shortfall between your work fingerprint and the requirements for those occupations makes explicit which of the 120 variables you are lacking in, and by how much. For example, the Anaesthesiologist occupation requires a minimum value of 6.46 in the Medicine and Dentistry knowledge category, which is near the top of human performance. Unless you are a specific kind of medical professional, you likely lack the knowledge to perform that occupation.

The O*NET work primitives are *latent variables*. Tests measuring one and only one of these 120 dimensions without entangling many of the others are difficult or impossible to devise. For example, testing knowledge categories requires a way to receive a question and return an answer, which requires ability categories.

Tests (devised by O*NET and others) to extract these latent variables in people assume many of the most difficult aspects of building an embodied intelligence are already in place. For example, taking a test designed to measure knowledge (for example, knowledge of law) requires many abilities we take for granted in people to be present, such as vision, ability to use a touch screen or a pencil; even the ability to recognize that a test is underway. Difficulties in measuring and comparing latent generators of functional abilities in non-human animals and machines (and also people) are well documented [(Hernandez-Orallo, 2017)](#).

When an AI system demonstrates some capability, such as passing law school exams [(Choi, 2023)](#), or playing Go [(Silver, 2017)](#), the issue of what we can infer about that system beyond the functional task they are demonstrating is a profound question. When people display such capabilities we naturally infer that this has some deeper meaning — that these are testing some latent properties that generalize to other settings. The test in itself may not be important; it's what it infers that we care about. In the case of law exams, the inference is that if you do well on those tests, you can do things lawyers actually do when they practice law. It is far from obvious that an AI system that aces a digital law exam could actually be a lawyer. A tool for use by lawyers, sure. But that is not what we are interested in here.

Tests of capability in people, from grade school tests through IQ testing to SATs to trades tests, measure very small differences between test takers that rests on a massive set of latent commonalities that make us human. It is largely these latent commonalities that we are interested in testing for and improving, and not scores on typical tests of intelligence or capability, even if those tests purport to measure latent characteristics. Our view is that working towards a system that can actually take a classroom test the same way a typical student would is a much better indicator of progress towards human-level AI than testing how well a digital system does on the test itself. This perspective has deep roots in the embodied AI and robotics communities [(Moravec, 1988)](#).

## 2.3. Inferring Robot Work Fingerprints Using a Portfolio of Subtask Fingerprints

How can we obtain the work fingerprint of a robot, given that it is formed of latent variables we cannot directly measure? We use a strategy of inferring these from test results on performance of a portfolio of *subtasks*. This procedure borrows both from the strategy O*NET uses to do the same with people [(ONET, 2023k)](#), and from the literature on devising psychometric (such as scholastic, personality individuation, or IQ) tests. That field is largely concerned with the objective measurement of latent constructs as well [(Kaplan, 2010)](#). Here is how this is done in our context.

We define a subtask to be any part of a work task or activity, including the entire task or activity. Subtasks can be much more concrete and granular than O*NET task descriptions; but they can also range up to the most abstract descriptions of work. In the context of the categories in Figure 1, any descriptor of work at any level can be a subtask, but so could activities not shown that are even more concrete. Marcus' list of five tests for AGI are each subtasks [(Marcus, 2022)](#), as are much more basic capabilities such as reading an eye chart, or being able to recognize and press a button. Because of issues related to Moravec's paradox, the hardest parts of building an embodied cognition are only captured indirectly by typical work task statements. We need a way to augment self-reported work task lists (such as those in Table 1) with more basic types of activities that those depend on and take for granted.

For any subtask, one can ask practitioners of that subtask to score which work primitives are required to do it successfully, and the level of capability in each that they view as minimum requirements. Questionnaires O*NET uses to do this can be found here [(ONET, 2023l)](#). This scoring approach creates an empirical self-reported map from subtasks to minimum values of work primitives required to perform the subtask in question. These minimum values are defined in the same space as a person's (or robot's) 120-dimensional work fingerprint. We call these minimum required values *subtask fingerprints*. Each subtask has one. In particular, each of the 19,265 O*NET tasks has one, as do each of the 2,068+334+41+9+4=2,456 work activities and each of the 1,015 occupations (see Figure 2 for examples, and the *O*NET Occupations Ranked by g+* tab in [(Data, 2023)](#) for all occupation fingerprints). This empirical mapping process can be used regardless of the nature or level of abstraction of the subtask and is not restricted to O*NET tasks, work activities, or occupations.

Full occupations typically require high levels of many of the latent work primitive values. These are interesting, but we also have a strong interest in much less ambitious subtasks. Shown in Figure 3 are example subtask fingerprints of some simple subtasks.

Given a subtask fingerprint, we can infer that if a person (or a robot) can successfully accomplish that subtask, their work fingerprint must be such that it possesses at least the minimum levels of work primitives required to perform the subtask. This then provides a straightforward way to infer the work fingerprint of an entity from whatever collection of subtasks the system can successfully accomplish.

We infer a work fingerprint from the performance of a portfolio of subtasks following the procedure defined in Algorithm 1.

**Algorithm 1: Extract work fingerprint from portfolio of subtasks**

1. Define a portfolio of subtasks, with each subtask including demonstrations and text descriptions
2. For each subtask, use people to empirically map it to its subtask fingerprint
3. Attempt to perform all subtasks in the portfolio
4. Over the set of successfully performed subtasks, compute the maximum value over that set for each of its 120 dimensions; those values define the inferred work fingerprint for the person attempting the subtask portfolio

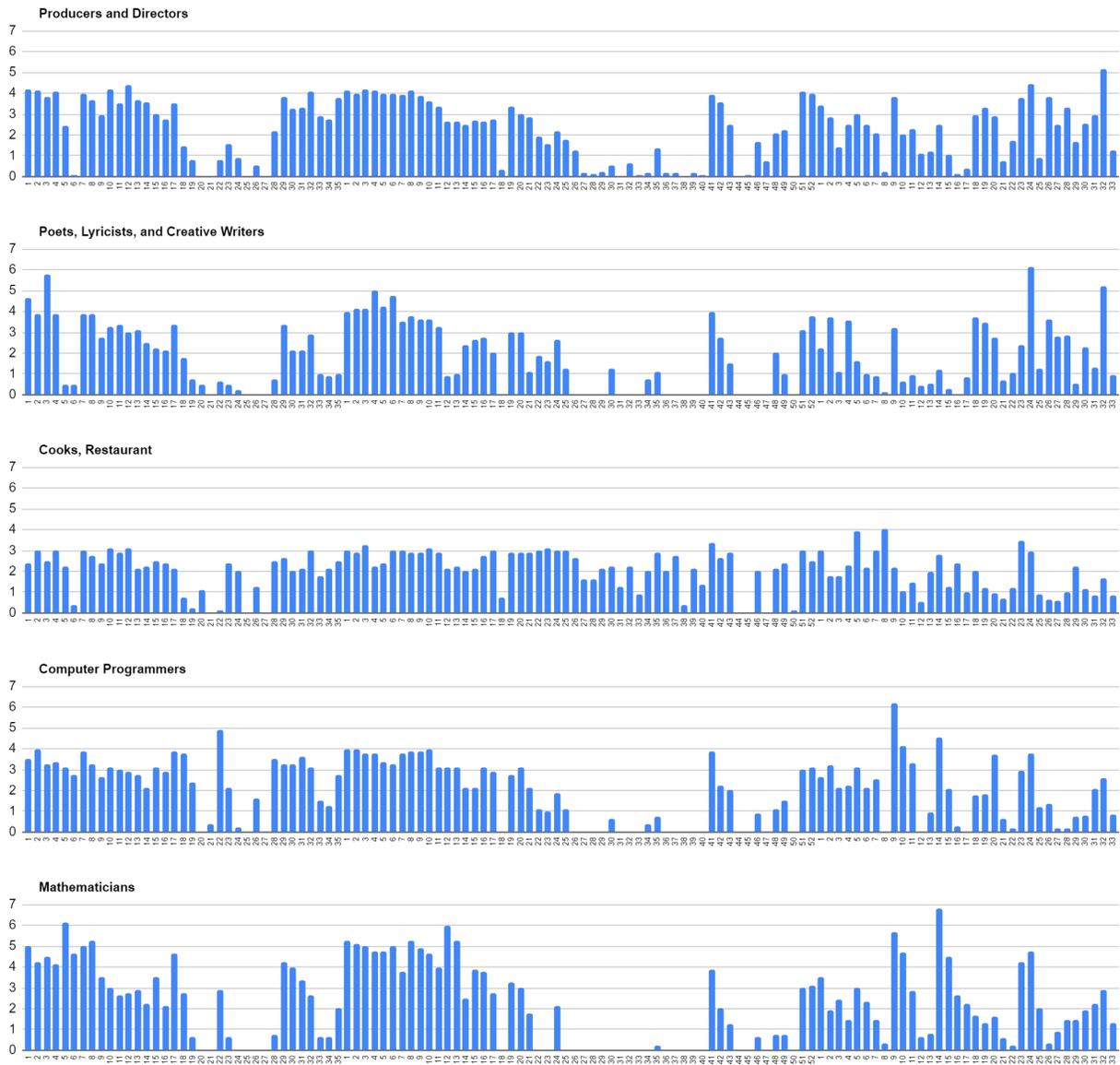

**Figure 2.** Subtask fingerprints for five sample O*NET occupations (choices inspired by a person (or machine) who would clearly pass Marcus' test for AGI (Marcus, 2022)). Ordering left to right on the horizontal axis are skills, abilities, and knowledge; numbering is as shown in (Data, 2023).

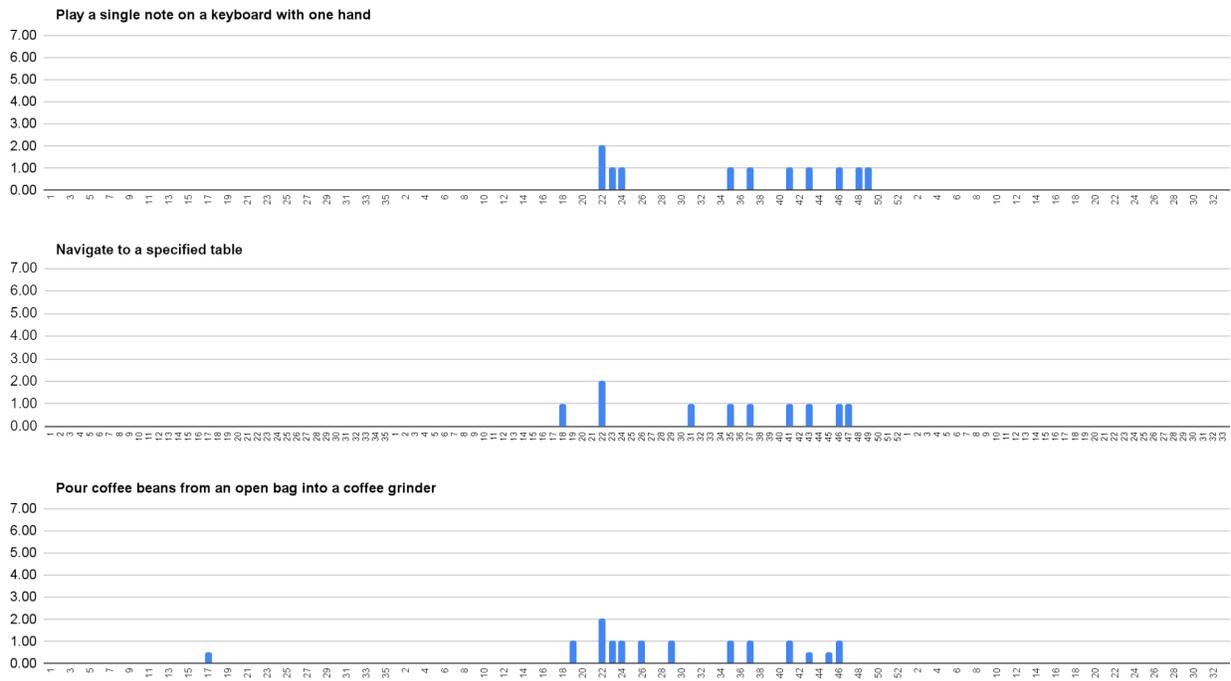

**Figure 3.** Example subtask fingerprints for three simple subtasks.

### 2.4. Inferring Unseen Subtask Performance From Work Fingerprints

Imagine a person has demonstrated that they can successfully perform all five occupation-level subtasks shown in Figure 2.[1] This person would have been employed as a Producer and/or Director; a Poet, Lyricist and/or Creative Writer; a Restaurant Cook; a Computer Programmer; and a Mathematician. This would be a very accomplished person with a large range of capabilities. Using Algorithm 1, we can extract their work fingerprint, where the subtask portfolio is formed of those five occupations.

We would like to now infer, based on that work fingerprint, what other subtasks could be performed, without actually performing those other subtasks. We do that by using the following simple procedure.

Given an unseen subtask, we consider it *performable* if each dimension of the subtask fingerprint is less than the same dimension in the person's work fingerprint.

To see a specific example, let's see if we can infer that the person above could perform the Models occupation. Both the Models occupation subtask (blue) and person's work (red) fingerprints are shown in Figure 4. Interestingly, we see that we cannot make that inference. The Models occupation, which has the lowest overall requirements of any O*NET occupation, has dimensions that have higher requirements than those demonstrated by performance of the five occupations in Figure 2. Therefore in our terminology, the Models subtask is not performable given this work fingerprint.

For humanoid robots, it will be a while before we can use full occupation automation as subtasks to find their work fingerprint. Instead we have to use much less ambitious subtasks, for which the minimum requirements to perform them are much lower than for full occupations. But the idea remains the same. We ask a humanoid robot to perform a portfolio of (much simpler and more concrete) subtasks, each of which has a subtask fingerprint, and then we extract the robot's work fingerprint using Algorithm 1.

---

[1] This would mean that this person would pass all five of Marcus' tests for AGI, as average practitioners in each of these occupations would excel at those tests.

Having extracted a robot's work fingerprint from its performance on a subtask portfolio, we can match this fingerprint against the subtask fingerprints for other as yet unseen work tasks to determine which we can infer can be performed, which can't, and (importantly) which dimensions have to change and by how much to unlock tasks that can't be performed yet. This allows us to infer what part of the US economy could be automated by a humanoid robot as a function of their work fingerprint.

In the limit, if a work fingerprint could be achieved with scores of 7 for all 120 dimensions, the inference would be that this robot could perform all occupations in the US economy, as its work fingerprint would encompass all occupation subtask fingerprints.

### 2.5. The g+ Score

A work fingerprint is a 120-dimensional vector of integers, each in the range 0 to 7. While this is the fundamental vector we track, we additionally define a scalar that is useful for tracking progress over time. We call this scalar g+ (pronounced g plus).[2]

To compute g+, first note that as we have subtask fingerprints for all occupations in O*NET, it is straightforward to first compute the sum over all 120 dimensions for each of these (giving a number between 0 and 120*7 = 840) and then take the average over all 1,015 occupations (doing so results in a mean value of 267.3).

We then define g+ to be the sum of all 120 scores in either a work or subtask fingerprint multiplied by 100/267.3. This normalization makes it so that a score of 100 is equal to the average value across all occupations. g+ ranges from 0 to 314.3. The standard deviation of minimum g+ scores required for occupations is 15.4 (see *O*NET Occupations Ranked by g+* in (Data, 2023)). This makes the distribution of g+ required to perform occupations very similar to IQ, which has a defined mean of 100 and standard deviation of 15.

The subtask g+ scores for Producers and Directors; Poets, Lyricists and Creative Writers; Cooks, Restaurant; Computer Programmers; Mathematicians; and Models, are 102.0, 84.5, 91.3, 91.7, 100.9, and 44.7 respectively. The Model g+ score of 44.7 is the lowest across all O*NET occupations. This means that a person or humanoid robot with a g+ score of less than 44.7 is not capable of performing any full occupation, regardless of distribution of strengths and weaknesses.

A person who successfully performs all five of the Producers and Directors, Poets, Lyricists and Creative Writers, Cooks, Restaurant, Computer Programmers, and Mathematicians subtasks would have a g+ score of 143.9 (but note, would still not qualify for the Model role even though its required g+ is much lower — see Figure 4).

---

[2] The notation g+ is derivative of the 'g' concept from psychology (Wikipedia 2023).

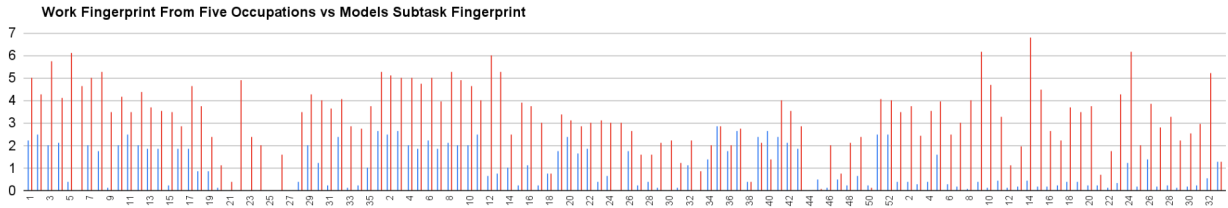

**Figure 4.** An individual who successfully performed all five Producers and Directors, Poets, Lyricists and Creative Writers, Cooks, Restaurant, Computer Programmers, and Mathematicians occupation subtasks would have the work fingerprint shown in red, formed by taking the max over minimum required values for each dimension of subtasks performed. We can then compare this work fingerprint to unseen subtask minimum values to infer whether or not the person could perform that new subtask, and if not, where they are lacking. In this case we compare to the Models occupation, and see small shortfalls in ability categories 39 (Gross Body Coordination), 40 (Gross Body Equilibrium), 45 (Peripheral Vision) and 50 (Sound Localization). Our approach therefore infers that the Models occupation subtask is not performable.

g+ is useful for tracking progress over time, but it comes with the following important caveats:

1. We are making the choice to value all work primitives equally in the g+ score. It is true that not all of them contribute in the same way to the types of work a person can do, but it is unclear whether adding in these weighting factors at this point helps us much. This would change the details of the metric, but not the general concept.

2. Reducing a 120-dimensional work fingerprint to a single number obscures important information about specifically what a person or robot with a certain g+ score can do. Two people with the same g+ score can have dramatically different strengths and weaknesses. For example, one could be very physically capable and the other could be very cognitively capable and they could score the same. The example used in Figure 4 illustrates this issue.

3. We make the assumption that the analysis in O*NET, assuming these tests are for people, translates to machines, in the sense that a robot that scored highly on the work primitives would be good at performing the tasks that underlie occupations. In Section 5 we test this assumption.

## 3. A Humanoid Robotic Embodiment

Our primary goal is to build a human-like egocentric mind. This mind is implemented as a software-based control system for a humanoid robot whose goals are to excel at work, in the sense of being able to perform most or all of the O*NET occupations described in Section 2.

Our perspective is that the physical robot is a tool for the use of a human-like mind to achieve the kinds of goals the mind favors. The physical robot's purpose is to be a vehicle through which the mind can sense and act on the world to achieve its goals. We would like to build physical robots that, if our own minds were immersed in them, would allow us to sense and act on the world much like we do in our own bodies. The mind and its goals are primary; the body is instrumental.

A humanoid robot designed to be capable of performing general tasks in the economy will likely perform worse than special purpose technologies on the tasks those special purpose technologies are designed for. However, just like people, humanoid robots should be able to use special purpose technologies in the same way we do when appropriate (including powerful weak AI systems), and therefore subsume their capabilities.

### 3.1 Embodiment Specification

Here we present an example of the type of humanoid robot we have tested using the methodologies introduced in Section 2. We refer to it as the Sanctuary Fifth Generation General Purpose Robot, or GPR5. When necessary to distinguish individual robots of this type, we will use the nomenclature GPR5-N, where N refers to the specific robot. All of the tests reported on in Section 5 were performed using GPR5 robots.

GPR5 comprises a humanoid torso, arms, hands, neck, and head, with a wheeled base providing limited mobility. The system requires a tether for power and networking (see Figure 5 and (Video, 2023)).

Shown in Figure 6 are the specifications of the torso, arms, neck, and head subsystems. From the perspective of information flowing into and out of the torso, neck, and head, there are 17 actuator degrees of freedom, two types of camera (Kinect and Zed Mini), a pair of stereo microphones, and one speaker.

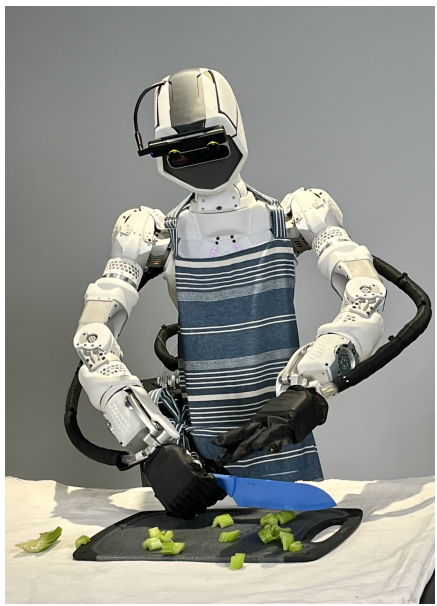

Figure 7 provides specifications on the hands. Shown are the specs for the Sanctuary World's Best Hand (WBH) v1.75. These hands are hydraulically actuated, with 16 degrees of freedom per hand. Each hand has 19 pressure sensors monitoring hydraulic pressures, and 480 Force Sensitive Resistor (FSR) haptic sensors distributed over the fingers, thumb, and palm.

Figure 8 provides the specifications of the wheeled base. Bipedal locomotion systems are under development (Ackerman, 2023) but these are not yet mature enough to be integrated into the full system.

**Figure 5.** GPR5-4 chopping celery. This is a subtask related to food preparation type tasks.

| GPR5 Torso, Arms, Neck, and Head | | | |
|---|---|---|---|
| Subsystems | Torso | Arm | Head + Neck |
| Total Weight (kg) | | 16 | |
| DOFs | 2 | 6 | 3 |
| Total DOFs | | 17 | |
| Arms Length Payload (kg) | | 0.5 | |
| Control Mode | | Impedance Control | |
| Actuators | | Electromechanical | |
| Actuator Sizes | RCR150-100 | RC60, RC135 | S14 |
| Cameras | | | Kinect and ZMini |
| Communication Protocol | | EtherCAT | |
| Joint Work Space (deg) | Torso Flexion (-15,+30) | Shoulder Flexion (-45,+180) | Neck Abduction (-30,+30) |
| | Torso Rotation (-45,+45) | Shoulder Abduction (0,+120) | Neck Flexion (-45,+70) |
| | | Shoulder Rotation (-90,+90) | Neck Rotation (-90,+90) |
| | | Elbow Flexion (-10,+150) | |
| | | Forearm Rotation (-90,+90) | |
| | | Wrist Flexion (+60,-45) | |

**Figure 6.** GPR5 torso, arms, neck, and head specifications.

| WBH V1.75 | | | |
|---|---|---|---|
| Subsystems | Finger | Thumb | Palm |
| Weight (kg) | | 1.2 | |
| DOFs | 3 | 4 | 0 |
| Total DOFs | | 16 | |
| Payload (kg) | | <1 | |
| Speed (deg/s) | | ~150 | |
| Control mode | | Position Control With Force Cap | |
| Actuators | | Hydraulic With EPOS4 and electrically driven pistons | |
| Hydraulic Pressure (psi) | | 400 | |
| Max Force (N) | 10 flexion, 5 abduction | 20 flexion, 20 abduction | |
| Positional Accuracy (deg) | | 1-2 | |
| Encoder Type | | Absolute encoders at motor, no encoders at joint | |
| Pressure Sensors | | ForceN sensors on valve pack | |
| # Sensors | | 19 | |
| Sample Rate (Hz) | | 100 | |
| Haptic Sensors | | FSR 1 DOF | |
| # Sensors | 16 (DP),16 (IP), 16 (PP) | 16 (DP), 16 (PP) | 256 |
| Total Haptic Sensors | | 480 | |
| Sample Rate (Hz) | | 100 | |
| Communication protocol | | EtherCAT, CAN | |
| Joint work space (deg) | MCP Flexion (0,+90) | Thumb Opposition (0,+90) | |
| | MCP Abduction (-15,+15) | Thumb Abduction (0,+60) | |
| | Coupled PIP (0,+100) | CMC Flexion (-15,+60) | |
| | Coupled DIP (0,+85) | IP Flexion (0,90) | |

**Figure 7.** Sanctuary World's Best Hand v1.75 specifications.

| Mobile Base V1.0 | |
|---|---|
| Weight (kg) | 25 |
| DOFs | 2 |
| Actuator | Segway RMP 220 Lite |
| # Actuators | 2 |
| Motion Type | ZTR rotation, forwards and backwards |
| Sensors | Wheel Encoders, IMU |

**Figure 8.** Wheeled base specifications.

## 3.2  Sensor and Actuator Data Flow

In Section 4 we will provide an overview of the control system for this type of robot. In keeping with the philosophy that the robot is a tool for the mind to accomplish its goals, here we explicitly characterize the information flowing into and out of the control system. This information flow defines the interface that the mind has to its environment.[3]

Sensor and actuator data for the system used here is shown in Figure 9. The control system (described in detail in Section 4), represented as the dark gray box on the top right, is a function that takes as input sensor data and outputs actuator data.

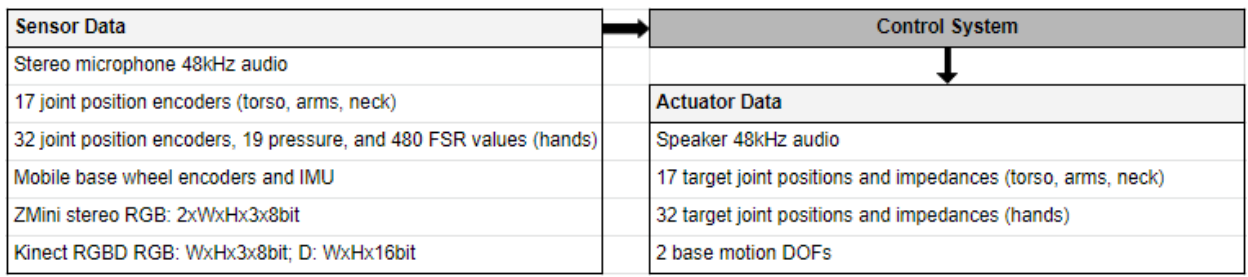

| Sensor Data | | Control System | |
|---|---|---|---|
| Stereo microphone 48kHz audio | | | |
| 17 joint position encoders (torso, arms, neck) | | **Actuator Data** | |
| 32 joint position encoders, 19 pressure, and 480 FSR values (hands) | | Speaker 48kHz audio | |
| Mobile base wheel encoders and IMU | | 17 target joint positions and impedances (torso, arms, neck) | |
| ZMini stereo RGB: 2xWxHx3x8bit | | 32 target joint positions and impedances (hands) | |
| Kinect RGBD RGB: WxHx3x8bit; D: WxHx16bit | | 2 base motion DOFs | |

**Figure 9.** Sensor and actuator data flow for the GPR5 system.

---

[3] This data flow picture omits closed loop control occurring at the level of the actuators themselves. We assume that impedance and position targets sent by the control system to the actuators are handled by a separate system that is responsible for reaching specified targets.

## 4. Carbon: An Autonomous Control System for Humanoid Robots

Carbon (a partial acronym of Cognitive Architecture for Robots) is a software-based autonomous control system for the type of humanoid robot described in Section 3. Our intent is for it to model and replicate the properties of the human mind. Its progress towards this long-term objective is measured via extraction of work fingerprints and associated g+ scores using the process described in Section 2, current and historical results of which we report on in Section 5.

### 4.1 High-Level Overview

Carbon is a *cognitive architecture*. Its role in the overall system is to convert sensor data into actuator data, as shown in Figure 9 (Carbon is the gray box labeled 'control system') and Figure 10 (where the overall context is added, but the sensor and actuator data are the same).

Cognitive architectures are 'architectures' in the sense that they propose dividing up the complexity of a mind into brain-inspired components that interrelate in some brain-inspired way (Kotseruba, 2020). This modularizes engineering tasks, and provides a natural way to separately define different components of the core system and learning paradigms described in Section 1.

Designing a cognitive architecture requires making difficult choices about how a mind should be structured. There are powerful and compelling arguments in favor of minimizing such choices (Sutton, 2019). We have to guess what the pieces of a mind are, and how they are interrelated, and our intuitions about how minds work are unreliable (Dennett, 1998).

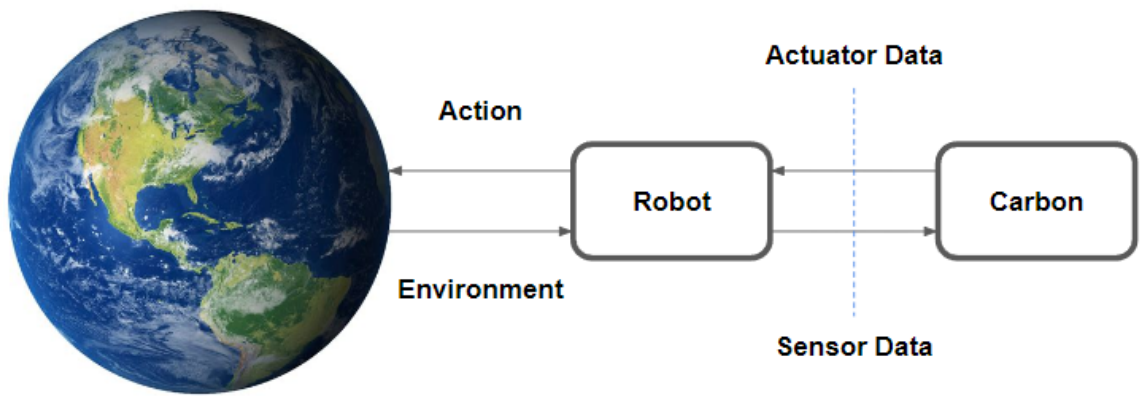

**Figure 10.** A robot's sensors convert information from the environment into sensor data, which is the input to the Carbon system. Carbon's output is actuator data — the signals sent to the actuators in the robot, including audio output. Actuation causes changes in the environment, which then change the sensor data. Specific sensor and actuator data used here are shown in Figure 9. The dashed blue line shows where we define an 'information API' into and out of Carbon. In the usual reinforcement learning agent/environment interaction picture, sensor data are (generally non Markovian) observations.

Our view is that, at least at the time of writing, the benefits of using a structured approach outweighs its drawbacks. These benefits include modularization (modules can be developed independently with clear APIs); broad-strokes versions can be quickly brought online; existing high quality tools, such as [Unreal Engine](), [Cyc](), and [Large Language Models]() can be included in a straightforward way; improved scrutability and explainability; debugging is much easier; they are easier to match to cognitive science/neuroscience understanding (i.e. more bio-inspired); they are easier to transfer to new embodiment versions; and require fewer compute resources to train. Our future understanding and technologies may reach the point where we won't need the explicit structures we introduce here. But we are not there yet.

Carbon uses a design philosophy heavily favoring the use of priors. Much of what makes us human comes from our evolutionary heritage, and is not learned during one lifetime; or at least learning only plays a part. While there are aspects of this problem that are 'tremendously, irredeemably complex' [(Sutton, 2019)](), not everything is. We accept the possibility of bitter lessons in exchange for engineering clarity. If it turns out we have made bad design decisions, that will show up in our testing process, and we can change those decisions.

### 4.2   Analogous Teleoperation

We extend the architecture in Figure 10 to include an *analogous teleoperation* system. This system comprises a hardware system, which we call a *pilot rig*, designed to relay sensory signals coming from a remote humanoid robot to a person, who we call a *pilot*, and convert the person's movement (including speech) into actuation signals sent to the robot. The extended architecture is shown in Figure 11. Note that the input and output signals to Carbon and to the analogous teleoperation system are *identical*. The switch allows an external operator or software system to route the sensor and actuator data to / from either controller.

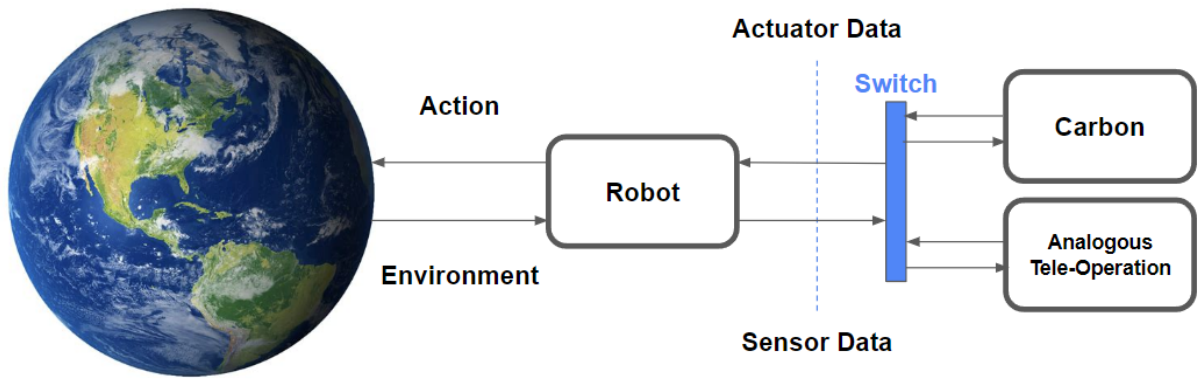

**Figure 11.** Sensor data can be sent to an analogous teleoperation system, where a person converts that information into actuator data. In this architecture the switch only changes the paths of the information flow, and not their type or content. Carbon and the analogous teleoperation system have identical input and output data.

#### 4.2.1 Pilot Rig Hardware

We require the target humanoid robot to analogously follow the movements of the human operator. For this to be possible, we must first have some way of tracking the motion of the human operator. We also require hardware that can facilitate the transmission of sensory information back to the human operator from the remote environment. Here we list important hardware components of the teleoperation system currently in use.

##### 4.2.1.1 HaptX Gloves

This system is responsible for tracking the motion of the human operator's fingers. These gloves also provide force feedback through skin deformation.

##### 4.2.1.2 Contoro Rig

The Contoro rig is an exoskeleton designed specifically for our teleoperation setup. Contoro uses encoder measurements, along with the forward kinematics of the device, to determine the position and orientation of the pilot's limbs. The Contoro rig has motors within its joints, allowing the device to provide force feedback to the pilot.

##### 4.2.1.3 Valve Index Headset

Visual and audio feedback is provided to the pilot using a Valve Index VR headset. This headset is the pilot's primary channel of information regarding the robot and its surrounding environment.

##### 4.2.1.4 Bidirectional Pedals

These pedals allow the pilot to control the robot's mobile base.

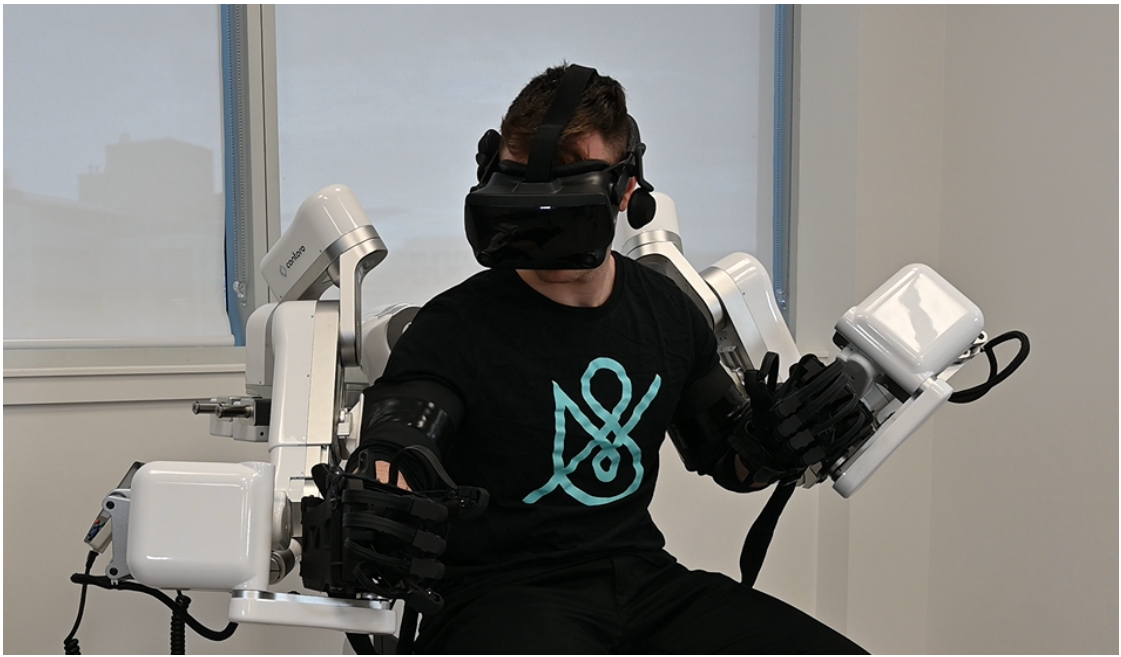

**Figure 12.** Pilot rig, including Haptx gloves, Contoro rig, Valve Index headset, and pilot. Not shown here are the bidirectional pedals for driving the base.

#### 4.2.2 Pilot Rig Software

The analogous teleoperation system is written within the [ROS 2](#) framework. Following ROS2 convention, different tasks are carried out by *nodes*. Nodes are executables that are responsible for carrying out a meaningful piece of work. Analogous teleoperation uses a ROS2 launch file to run all the required nodes for operation. Here is a list of ROS2 nodes used by the system.

- **Core Analogous Teleoperation (CAT):** The CAT node is responsible for tracking the pilot's movements. The CAT node receives position measurements from the Contoro node. A call is made to an inverse kinematics (IK) solver to determine the joint space commands that should be sent to the robot.
- **HaptX**: The HaptX node reads the finger positions and communicates these measurements to the CAT node. The CAT node uses these measurements as part of the argument to the IK solver.
- **Pedal:** The pedal node reads raw voltages from the pedals and maps them to appropriate linear and angular velocity commands which are sent to the mobile base. The node reads signals from the bidirectional pedals via serial communication.
- **Audio-Video:** The AV node is responsible for coordinating audio and visual signals between the robot and the pilot rig.
- **Contoro:** This node reads the measurements from the Contoro system. Positions are published in Cartesian space and consumed by the CAT node.
- **CAT GUI:** This node launches a GUI that provides visualization of the goal robot configuration (determined by IK solver) and the true robot configuration. This GUI also provides useful information regarding sample times for different sensors.
- **Base:** This node receives linear and angular velocity commands to move the base of the robot from the pilot station.

### 4.3 High Fidelity Simulations of Robots and Environments

In the previous section, we added an alternate pathway for converting sensory signals from the environment into actuation. Here we do something similar, except now we add an alternate *environment* — the left hand side of Figure 11 that produces the sensor data and consumes the actuator data.

We add a pathway for *digitally simulated* robots and environments, as shown in Figure 13. As was the case in the previous section where we added analogous teleoperation, we require that the data types flowing into and out of the physical robot are exactly the same as those flowing into and out of the simulated robot.

This choice enables four modes of operation, based on the choices of switch settings: Carbon controlling a real robot in the real world; Carbon controlling a simulated robot in a simulated world; analogous teleoperation of a real robot in the real world; and analogous teleoperation of a simulated robot in a simulated world.

Enabling a simulated robot in a simulated environment that is indistinguishable from a physical robot in the physical world to a controller (such as a pilot) is difficult. Our solution to this is a software system we call the Sanctuary World Engine, which is a tool for designing high fidelity physical simulations. World Engine's architecture is shown in Figure 14. An example first person perspective in both real and simulated environments is shown in the topmost row in Figure 15.

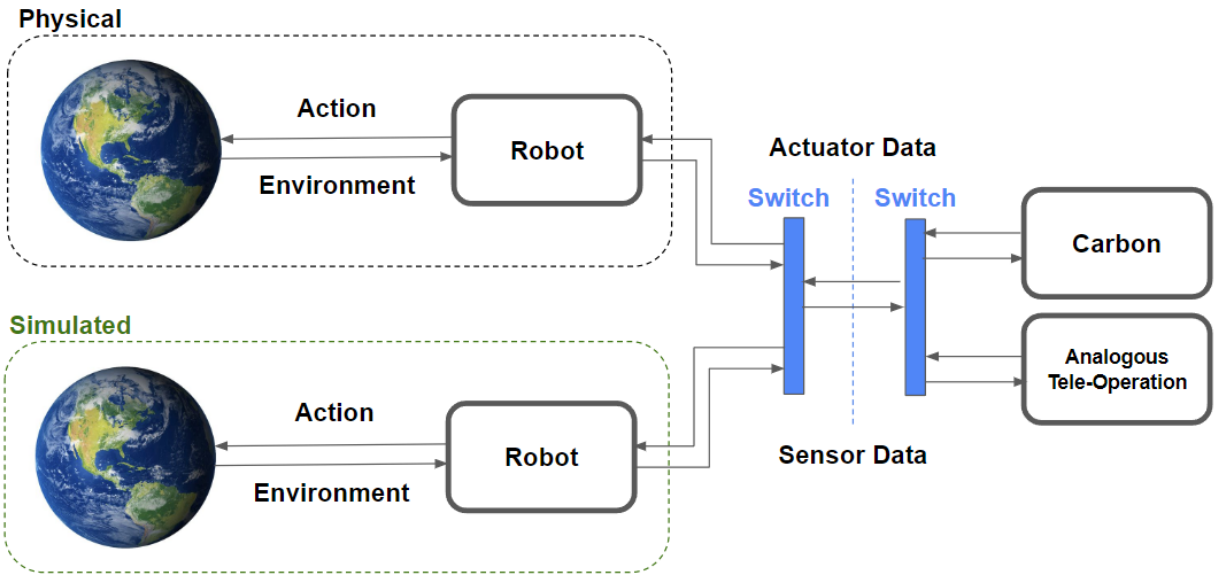

**Figure 13.** The data streams forming the input and output to Carbon or an analogous teleoperation system can arise from physical robots in real environments, or from simulated robots in simulated environments. The data types and formats forming actuator and sensor data are identical between real and simulated environments.

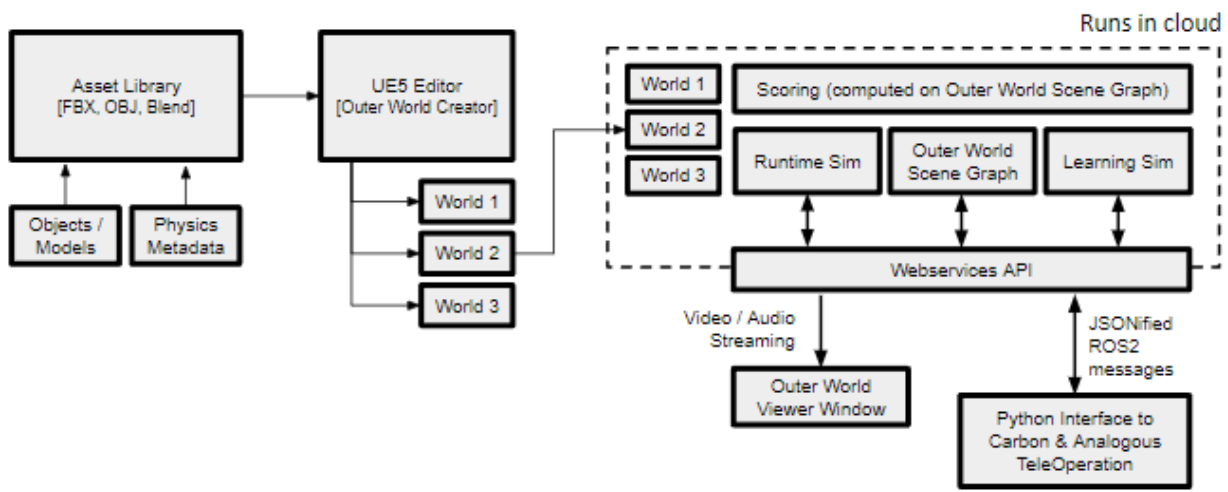

**Figure 14.** Architecture for the Sanctuary World Engine. Assets are created by artists in Blender, converted to a variety of useful file types, and stored in an Asset Library. Physics metadata is applied to them by hand at this point. These are then available for import into Unreal Engine 5 where specific Environments are defined (these are called Worlds in this diagram). Once these are complete, they are pushed to a cloud instance where they can be accessed via a web services API which exposes aspects of these Worlds, and in particular their scene graphs and the reward associated with these (computed in the Scoring module), a real time simulation (this is the use case in Figure 13) and a headless variant using Omni Gym that can be used to generate and use training examples for consumption by learning algorithms (labeled Learning Sim here). The 'outer world' terminology refers to the kind of simulation in Figure 13, meant to simulate the real world in real time. The JSONified ROS2 message passing system is responsible for managing sensor and actuator data flow between the digital simulation and the controller.

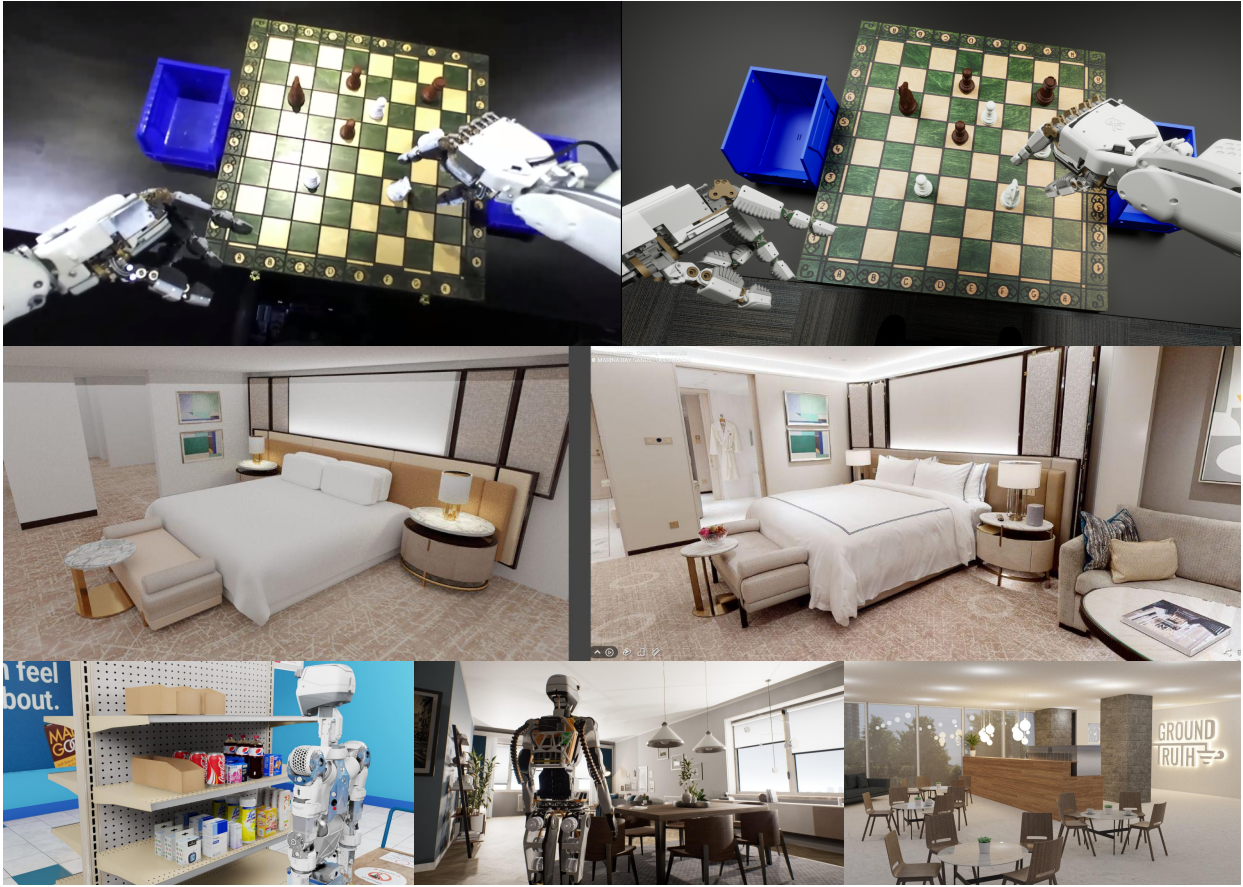

**Figure 15.** Top Left: snapshot of camera feed from a real robot in a real environment. Top Right: snapshot of simulated camera feed from a simulated robot in a simulated environment, created in World Engine. Middle left: simulated World Engine environment; middle right: physical bedroom. Bottom: a selection of simulated environments created in World Engine.

### 4.4   Ground Truth: An Experimental Environment for Subtask Testing

In Figure 13 there is a correspondence between real and simulated environments. Building such a correspondence, such that the controller can't tell the difference between the two (an example of a pilot's first person perspective into both is in the topmost row of Figure 15), is in general very difficult.

To lessen this difficulty we designed and built a real-world environment, which we call *Ground Truth*, and then used World Engine to build a high-fidelity simulation of it. This includes the creation of a library of assets matching all of the objects in the physical space. The digital variant of Ground Truth is a 'World' in Figure 14.

The physical Ground Truth is a 1,505 square foot environment located within the Sanctuary offices in Vancouver. It was designed to be both a functioning employee lounge, including coffee, food, comfortable seating, work tables, and a bathroom, and also a test environment within which the metrics and processes introduced in Section 2 can be run. The space contains a GPR5 system of the sort described in Section 3, which is tasked with performing all of the activities required to operate the space. This includes security, cleaning, operating the coffee and point of sale machines, stocking and organizing merchandise on shelves, and social interactions with customers.

The digital Ground Truth environment is as high fidelity a digital copy as we were able to make of the physical Ground Truth and its contents. Anecdotally we have seen that human pilots are not always able to initially distinguish between the physical and simulated Ground Truth environments. However as time proceeds, there are differences that arise, in particular in the physics of the interaction of objects with the robot's hands. These are 'tells' that allow a pilot to infer they are in the simulated environment. These differences have been systematically decreasing over time. There do not appear to be any fundamental obstacles to indistinguishability.

### 4.5 Carbon Architecture

We made explicit in Figure 13 that pilots and Carbon both convert a specific list of sensory data into a specific list of actuation data (listed in Figure 9). Any capabilities demonstrated by the robot under analogous teleoperation are delivered by the aspects of the pilot's mind captured by this picture. While we don't know how to characterize all of the signals coming into and out of a person's mind *in situ* inside the human body, we can (and have to) characterize all the signals passing to and from that person's mind to an analogously teleoperated robot outside of their body.

Carbon has the same job as the pilot's mind. If we call the input sensory data $X(t)$, the function the pilot's mind implements $M$, the Carbon system $C$, the output actuation data from the pilot $Y_M(t)$, and the output actuation data from Carbon $Y_C(t)$, the engineering problem is to create a function $C(X(t)) = Y_C(t)$ that is 'similar enough' to the function $M(X(t)) = Y_M(t)$, in the sense of being able to perform the same kinds of general work tasks the pilot can perform.

Building a reasonable C function encompasses both the original founding goal of the AI community and, to the extent human cognition can be considered general, aspects of the AGI problem. To make progress we use as much prior knowledge as possible about not only what C is expected to functionally do, but also how our own minds do these things. We wish to copy what is known about the internal structure of M into our design for C.

This strategy has a long history in the cognitive science community. Software systems designed to mimic how minds work from the perspective of internal modular structures and their inter-relationships are called cognitive architectures (Kotseruba, 2020). Examples of cognitive architectures include SOAR (Laird, 2019), ACT-R (Ritter, 2018), LIDA (Franklin, 2006), LeCun's 'path towards autonomous machine intelligence' (LeCun, 2022), OpenCog (Goertzel, 2014), and some reinforcement learning based systems (Sutton, 2018).

Carbon is a cognitive architecture. Its high level design has the following guiding principles:

1. **Copy what is known about how the human mind works**. The architecture draws inspiration from biology and neuroscience, specifically what is known about how our brains control our bodies. We explicitly include a first-person egocentric perspective, object-centric world models, task planning, logic-based common sense reasoning and inference, motion planning, methods for goal specification and quantification, and natural language generation and understanding.

2. **Understand human-like AI to be the same thing as a real-time embodied control system for a physical humanoid robot capable of general work tasks.** The architecture is designed to be run in real time on a physical robot with sensors and actuators providing enough richness to perform general work tasks of the sort described in Section 2.

3. **Design it to be an egocentric mind with its own agency**. We aim to build a mind that can solve many different problems across many different domains, and use tools including other weak AI systems and its own physical body to aid in goal-seeking behavior.

4. **Design it to be modular and technology agnostic, and progress it via evolutionary pressure driven by continuous incremental increases in work fitness as measured by g+**. Carbon's design is modular, allowing each module to use the right technology for its job. This includes state of the art connectionist and symbolic modules. Each module can evolve or be replaced with something better as technologies improve, as can the design of the overall architecture.

We start describing Carbon's internal structure by redrawing Figure 13 in a different way in Figure 16. Following Section 4.3 we call the kind of simulation introduced in Figure 13 the *outer world*. This is shown inside the shaded orange area. The outer world is a high fidelity simulation of reality, with identical data flowing into and out of both pilot rigs and the Carbon system.

We add colored dots where different types of memory are stored and used. These are labeled in the memory key in the cognitive architecture diagrams. Memory is always in the format and level of abstraction appropriate for the module accumulating it. Definitions of each memory type are provided in Section 4.5.2.16. Each place where memory resides is an accumulation point for the use of memories for different types of learning.

In Figure 16 Carbon remains a black box that takes in sensory data and outputs actuation data, from and into either a digital simulation (orange, labeled outer world) or physical reality (blue, labeled Physical Environment).

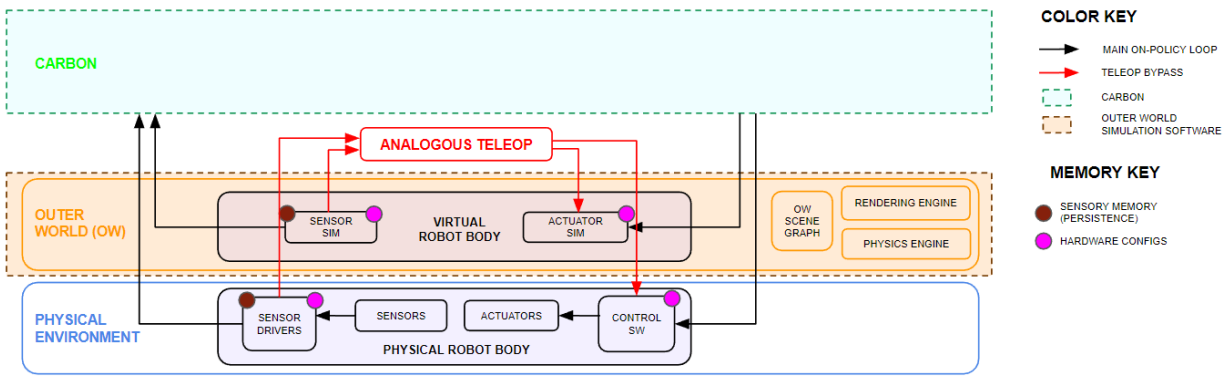

**Figure 16.** This is Figure 13 redrawn, with the addition of some detail and explicitly labeling where and what type of memory is used in each part of the architecture.

### 4.5.1 A Note on Timing

Carbon is designed to control a physical humanoid robot. This means that the system has to work in real time. The timing with which perception comes into the mind and action comes out of the mind is critical if the robot is to respond to changes in its environment quickly and appropriately.

Timing of the perception-action loop is not shown explicitly in the diagrams that follow, but the main loop itself must be able to run as fast as the fastest external events we want to respond to (which in our case is currently on the order of one pass through the loop every 100 milliseconds; this is roughly of the order of human response times [(Wikipedia, 2023a)](#)) to keep the robot fluidly responsive to environmental dynamics.

Timing is a very important part of the overall system design. There are many different timescales on which processes run within the mind, most of these much longer than the single loop pass time.

### 4.5.2 Looking Inside Carbon

Before we begin looking at the components of the Carbon system, we introduce a key underlying idea: a symbolic Instruction Set for digitizing action-taking.

#### 4.5.2.1 A Symbolic Instruction Set For Action Specification and Digitization

In the design of computer processors, an *Instruction Set* is defined which acts as a boundary between the sub-symbolic realm of physics (voltages and currents) and the symbolic realm of human-readable representations and semantics [(Hennessy, 2017)](#). The Instruction Set is typically the lowest level of abstraction available for programming a processor. Higher level languages are compiled down to sequences of these Instructions. The combinatorial explosion of possible orderings of Instructions contains within it all possible programs, in a similar way that all possible combinations of characters in the English alphabet contains all possible English writing.

We designed and implemented a conceptually similar system for physical action-taking, by defining a symbolic Instruction Set for the type of humanoid robot described in Section 3. Our current Instruction Set is included in the *Instruction Set* tab in [(Data, 2023)](#).

In our paradigm, all possible actions that can be performed by a humanoid robot are sequences of symbolic tokens, each of them corresponding to one of the allowed Instructions together with symbolic parameters. Each of the Instructions is implemented as a closed loop control algorithm, which we refer to as a *policy*, where success conditions are defined by Boolean functions we call *percepts*. We call the process of using policies to generate robot joint trajectories *motion planning*. Percepts are extracted from sensory data and are definitionally what it means to succeed in the execution of an Instruction.

Percepts are special functions of sensory data that measure (and define) whether or not the goal of an Instruction has been met. We conceptualize issuing an Instruction as being equivalent to specifying a goal of having its associated percept be true. Policies can be learned (for example, using reinforcement learning), or can be engineered.

For example, consider the Instruction [set_self_look_at_object($OBJ_UID)](#). The semantic meaning of this Instruction is that we ask the robot to center its visual gaze on a specific object that the system has assigned a particular identification integer to (for example, $OBJ_UID = 5 in Figure 17).

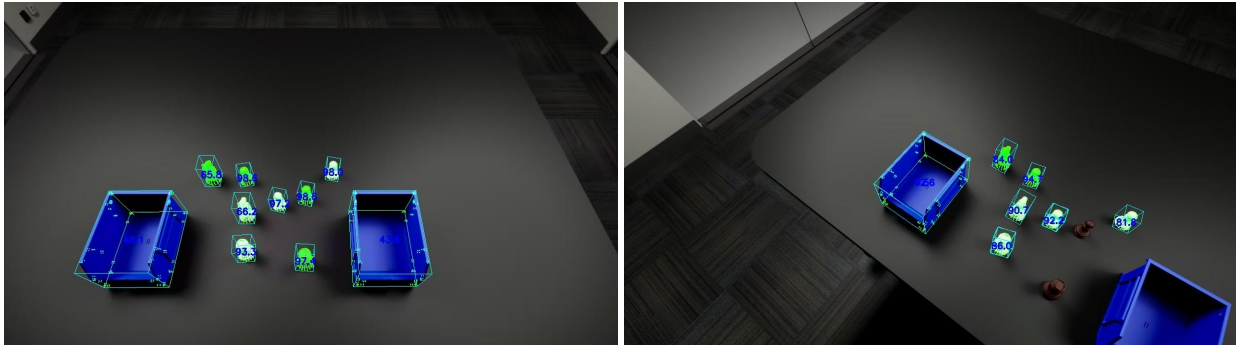

**Figure 17.** These images are from the first person perspective of the robot's cameras. Assume the system has assigned the leftmost blue bin $OBJ_UID=5. Left: the percept is_self_looking_at_object(5) associated with set_self_look_at_object(5) is false (the center of the object is too far from the center of the field of view); the policy is a closed loop control algorithm that moves the actuators in the robot towards centering the object. Right: is_self_looking_at_object(5) is true; the Instruction set_self_look_at_object(5) is reported to be successfully executed.

This Instruction sets a system goal of having its associated percept be true, and follows an engineered policy to achieve this goal.

The percept in this case is a Boolean function is_self_looking_at_object($OBJ_UID) that returns true if the robot's visual system has the object identified as $OBJ_UID centered in its field of view with some tolerance. The policy followed to achieve this goal in this case is a closed-loop control algorithm that uses incoming visual data to reduce the distance between the center of the camera field of view to coincide with the center of mass of the object.

The design strategy of creating a boundary between software (symbols) and hardware/physics (analog currents and voltages) in the case of computer processors has been very effective, as it allows high level programming languages to be abstracted away from the details of how the underlying Instructions are implemented. In action-taking, we have the vexing complication that there is no obvious analog to the transistor, or more abstractly the bit, which digitizes the underlying analog physics. In our Instruction Set implementation, the use of closed loop control policies with percepts that check whether actions have resulted in their designed effects is our proposed 'transistor analog'. Percepts are a way to digitize the fundamentally analog nature of action taking.

In a computer processor, an Instruction might ask that a specific bit in a memory register be set to 0. When that Instruction executes, analog currents and voltages flow in the physical structures that comprise the processor, but the final result is a symbolic outcome (the bit in question is either 0 or 1). In our architecture, when a command like set_self_look_at_object($OBJ_UID) is issued, just like in a processor, analog physics occurs in the actuators and sensory data, but the final result is a symbolic outcome, just like it is in a digital system (the percept is_self_looking_at_object($OBJ_UID) is either 0 or 1).

This pattern is required of all Instructions in our Instruction Set. They all come with clear human-readable semantic meanings; they are all specified entirely symbolically, including their parameters; each comes with a percept, which is a Boolean function of sensory data; each comes with a policy whose goal is making the percept true; executing this policy is also called motion planning. This pattern digitizes action-taking, allowing controllers to compile higher level task plans down to the Instruction Set in the same way conventional compilers do in computer processors.

With our Instruction Set defined, we now start looking at Carbon's internal structure. Our strategy in describing new modules will be to describe what they are for, how we implement them, and when feasible provide specific illustrative examples. We begin by describing the modular pieces added in Figure 18.

#### 4.5.2.2 Feature Extraction

Feature extraction takes low-level, sub-symbolic, high bandwidth, complex sensory data streams, including raw video, audio, proprioceptive and haptic streams, and converts them into high-level, symbolic, semantically meaningful information, which we call *features*.

Feature extraction is the first step in the process that Carbon uses to understand the world. Carbon's architecture allows arbitrary addition and removal of feature extractor functions that act on the sensor data. Each feature extractor is a specialized submodule that can be turned on or off by an attention module, which we introduce later on.

Feature extractors generally don't run all the time. The majority of feature extractors only run when needed. Some feature extractors do run all the time and publish their extraction results continuously.

Which feature extractors are included in the system is a decision made by system architects subject to the evolutionary pressure of gradually increasing fitness for work as measured by g+ scores. Each feature extractor module is either coded manually or learned from data to provide the desired feature extraction behavior. The system is designed to be able to easily add and remove third party feature extraction APIs (for example, the AWS, Google and Azure libraries).

Specifying features we think are relevant to downstream uses in Carbon is a form of introducing priors of the sort described in Section 1. While this restricts downstream uses to the features we extract, the system architecture is flexible enough to add or subtract features and therefore we can adapt this module if it is causing issues for system performance. We have found that if a certain feature type is required to enable some goal seeking behavior, it is usually obvious what is missing and straightforward to add.

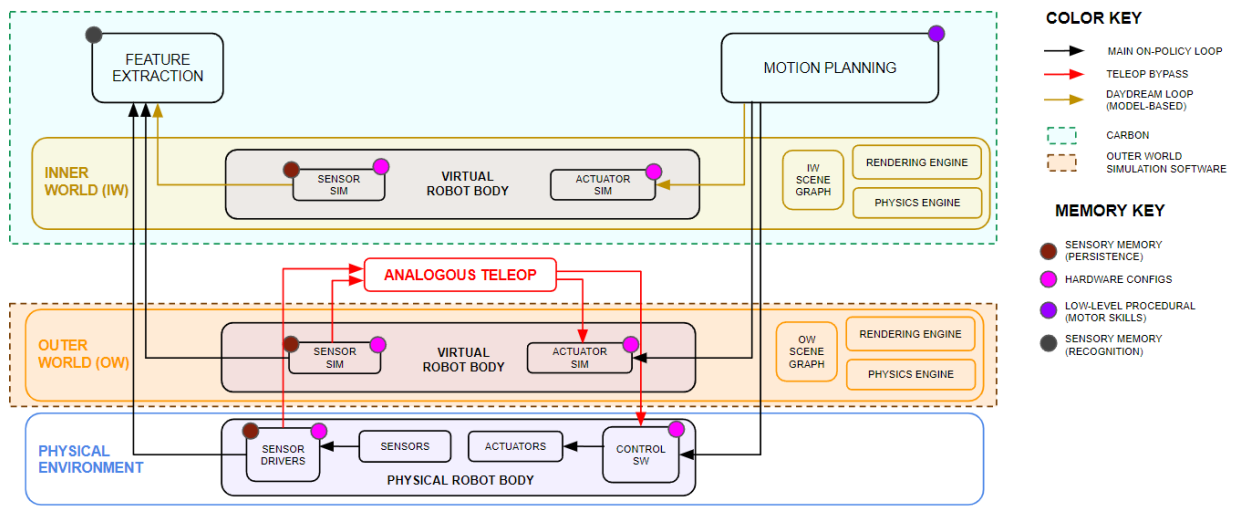

**Figure 18.** Added Feature Extraction, Motion Planning, and inner world model.

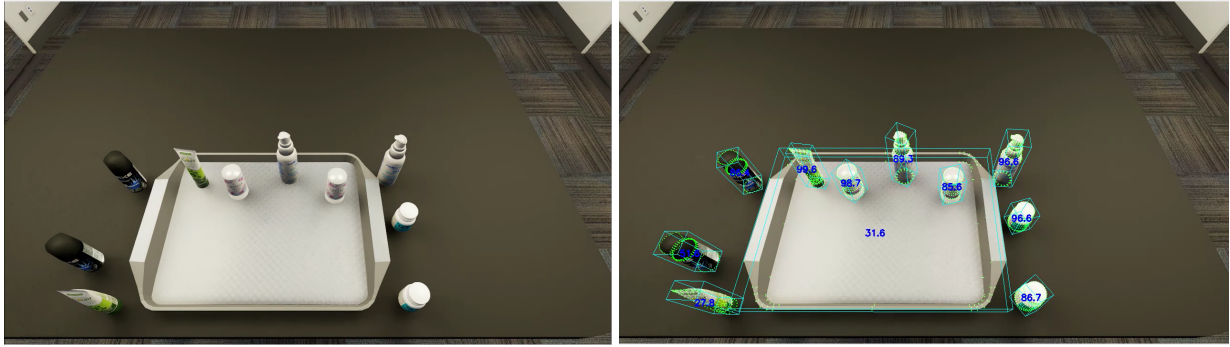

**Figure 19.** Left: feed from simulated robot cameras in a digital simulation (an outer world model). Right: extraction of positions, orientations, and object labels (labels not shown) from a portfolio of Carbon feature extraction modules.

Examples of some of the feature extractors that are implemented in the feature extraction module include:

- All the percepts defined in the *Instruction Set* tab in (Data, 2023)
- The location, orientation and labels of all detected objects (inverse graphics — see Figure 19)
- Mapping of the environment (multi-resolution SLAM)
- Text extracted from speech (speech to text)
- Text extracted from OCR on visual feed (reading text)
- Name labels extracted from recognizing specific people (facial recognition)
- The presence or absence of a hand in the scene
- Joint states for all actuators in the robot
- Number of faces in the field of view

The Carbon architecture allows arbitrary changes to this feature set.

### 4.5.2.3  Motion Planning

In Carbon, motion planning is the problem of figuring out how to move the robot's actuators to achieve an Instruction Set goal. The motion planner takes in an Instruction and converts it into low-level control signals, following the policy implemented for that Instruction. Note that motion planning in Carbon is restricted to serially implementing the closed-loop policies underlying the Instruction Set commands.

Carbon uses several different strategies for motion planning, based on the specific demands of the Instruction being implemented. These include optimization algorithms, learned RL policies, geometric motion planners (Cheng, 2021), and the RRT Connect algorithm (Kuffner, 2019).

### 4.5.2.4  An Object-Centric Inner World Model

In Figure 18 we added a pale yellow box inside Carbon, labeled the *inner world*, that has all of the same parts as the outer world. The inner world model is the egocentric model that a robot has of its environment. It uses the *exact same machinery* as the outer world simulation, including having a model of its own robot body from its own perspective. The contents of the inner world model are dynamically generated from the incoming sensor data stream.

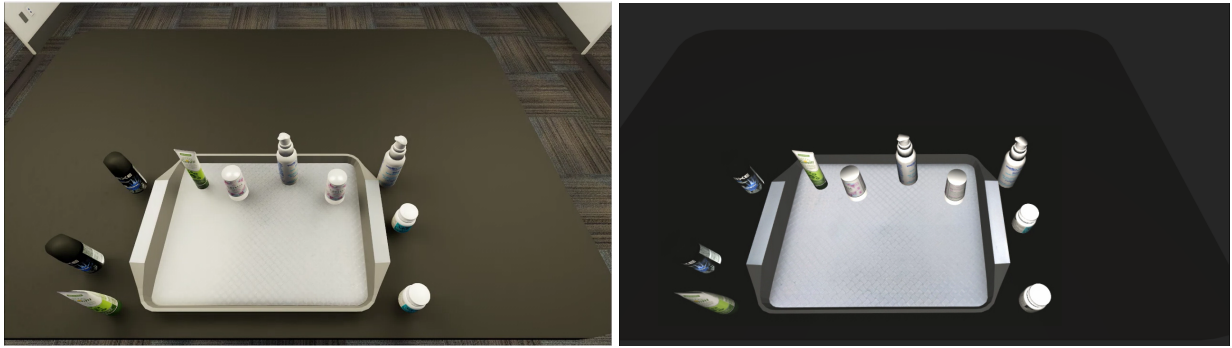

**Figure 20.** Left: same feed from simulated robot cameras in a digital simulation (the outer world model) as in Figure 19. Right: Carbon's inner world model, constructed from features extracted in Figure 19.

The inner world model is Carbon's egocentric model of its environment, including its own body. The information in the inner world model is incomplete. There is much more information in the outer world (in both physical reality and digital simulations of it) than is available to the system's own mind. This is the same as with human cognition. There are many facts about your current environment that are not within your own inner world model. Even your own body contains a multitude of truths of which your cognition is unaware or only partially aware, such as the specific locations of all the red blood cells in your body.

Sometimes the inner world model may be wrong, in the sense that it does not agree with the outer world model. In the example above, prior to 1667 and the discovery of the cell by Robert Hooke, the conceptions any human cognition had of their own blood could not have comported with our current understanding of reality. More prosaically, a robot may think an object is at a specific location and orientation, but the object is in a different location and orientation in the outer world.

Another common issue relates to object permanence. Is it difficult to correctly and continually ascribe the same $OBJ_UID to the same concrete object if a cognition's perception is not continuously aware of it. For example in Figure 20, if the robot were to exit the room and return, it is difficult to know whether the objects in the field of view are the same objects. Potential mismatches between inner and outer worlds are common and must be factored into the overall design of the cognitive architecture.

Cognitive scientists are largely united in the view that human cognition is object-centric. We build models of our environments focused on objects (including the robot's own body and its parts), relationships between objects, and object-focused temporal events (such as collisions between objects) (Dawson, 2013). This view is supported by growing evidence from advances in the realism of video games, which are explicitly object-centric by construction. Because of this, in both inner and digital outer world models, Carbon builds environments out of the composition of objects (including object-focused temporal events). All objects are modeled as shown in the Assets Library module of Figure 14.

In Figure 19, we connect sensory data coming from the inner world to the feature extraction module, and send actuator data from motion planning to the inner world. This allows the Carbon mind to imagine what the effects of action-taking would be inside its own model of the world before having to commit to actually taking those actions. This machinery allows Carbon to 'think about the world', and predict what will happen in the future subject to its taking actions at the sub-symbolic level.

We impose the constraint that the sensory and actuation data types be identical in all three of (1) physical reality; (2) the outer world model (a simulation of physical reality outside of the robot's mind); and (3) the inner world model (the robot's personal inner simulation of the environment it is in, which includes the simulation of its own body).

We now further expand the modules included in Carbon as shown in Figure 21.

### 4.5.2.5 System 1 Policy

The System 1 policy module comprises all policies associated with the execution of the Instruction Set elements.

### 4.5.2.6 Sub-Symbolic / Symbolic Boundary

This boundary separates sub-symbolic parts of the system (below the line) from symbolic parts of the system (above the line).

### 4.5.2.7 Concrete State Representation Updater (SRU)

As in reinforcement learning, *state representation* in Carbon comprises the system's full egocentric Markovian representation of the state of its environment. Note that state representation and the inner world model, while both models of the outer world, have different roles in the architecture. State representation is a representation of the current understanding Carbon has about the world. The system uses it to logically reason about and plan action sequences, both at the motion and task planning levels. The inner world model is a physics-based prediction engine that allows the system to model the sub-symbolic dynamical effects of action-taking on the environment. State representation is data; the inner world model is a dynamical prediction process.

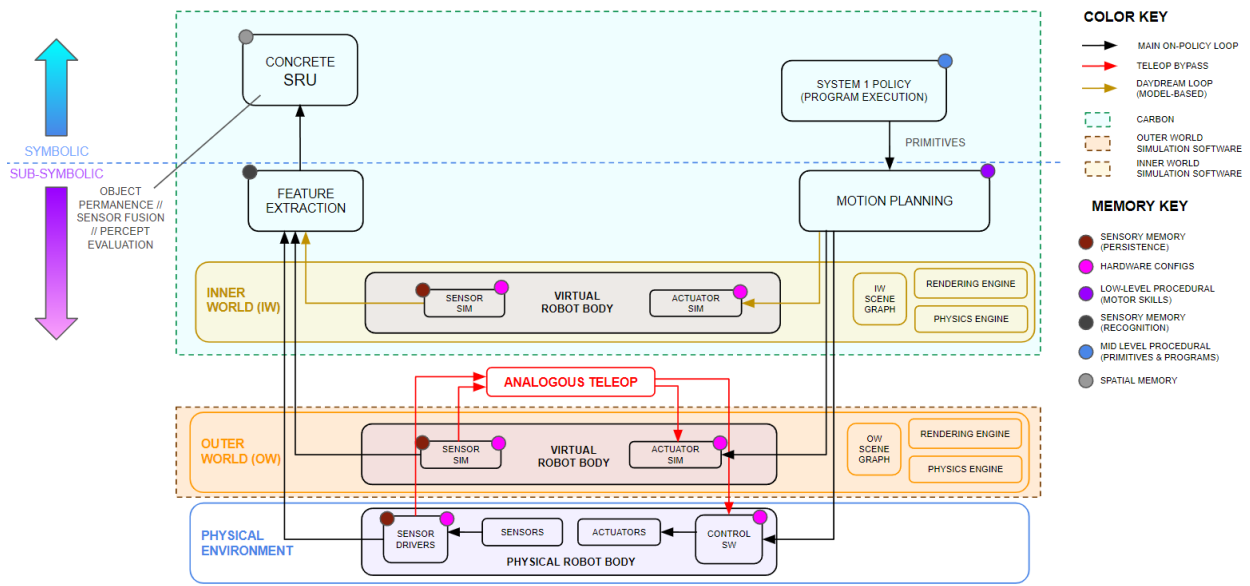

**Figure 21.** Added Concrete SRU, System 1 Policy, sub-symbolic / symbolic boundary, Instruction Set primitives.

In Carbon, the state representation contains three parts. The first of these, which we call the *concrete state representation*, is the part of the state representation derived from historical trajectories of the symbolic features extracted from sensory data in the feature extractor. Note that this also includes which actions from the Instruction Set were taken in what order, as this information is recorded by the feature extractor. We store these sequences in spatial memory (see Section 4.5.2.16 for memory types), which is a record of concrete symbolic properties of specific scenes the mind has experienced. An example of a part of the concrete state representation are the representations generated by multi-resolution SLAM algorithms, which generate persistent maps of the objects in the robots' environments.

The concrete state representation updater's role is to take the incoming feature stream and update the concrete state representation.

We now add new features shown in Figure 22.

### 4.5.2.8 Boundary Between "Thinking Fast" System 1 and "Thinking Slow" System 2

We introduce another boundary, this time between processes responsible for executing individual Instructions, which we call thinking fast, and processes responsible for deriving sequences of Instructions, also known as task planning, which we call thinking slow. These terms are (loosely) inspired by Kahneman's classifications (Kahneman, 2011).

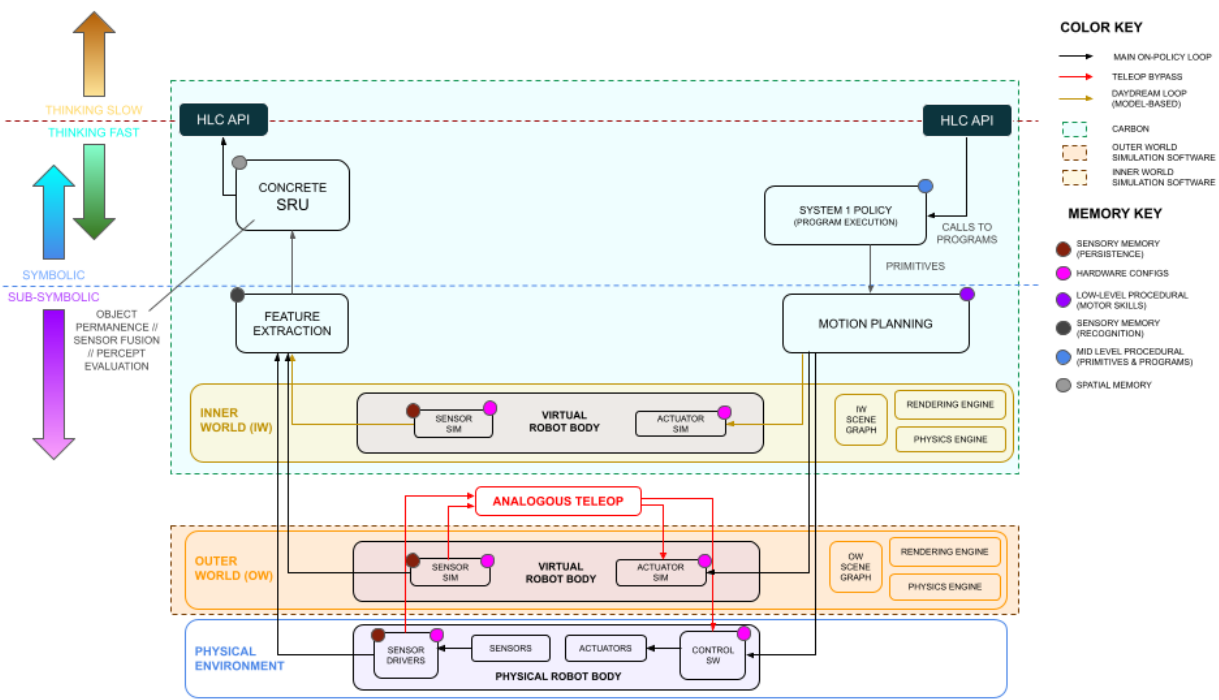

**Figure 22.** Added High Level Controller (HLC) API, thinking fast / slow boundary, calls to programs.

#### 4.5.2.9 High Level Controller API

The Carbon architecture accesses all system 2 processes via a cloud API, called the High Level Controller (HLC) API, that receives information from the concrete SRU and posts task plans in the form of serial Instructions sent to the System 1 Policy module.

We now add new features shown in Figure 23.

#### 4.5.2.10 State Representation

Everything within the pink outline is part of the Carbon system's state representation.

#### 4.5.2.11 The Thinking Slow Pathway

The dark blue shaded region contains modules that work together to convert Carbon's concrete state representation into a *task plan*. In Carbon, a task plan is defined as a serial set of Instructions. In the analogy to processors, a task plan is a compiled program. The modules within the dark blue shaded area are hosted in a cloud environment, remote to the robot embodiment. The API that connects these resources to a running Carbon instance is architected as a web API.

Carbon asynchronously posts the concrete state representation to the API, and asynchronously receives task plans from it.

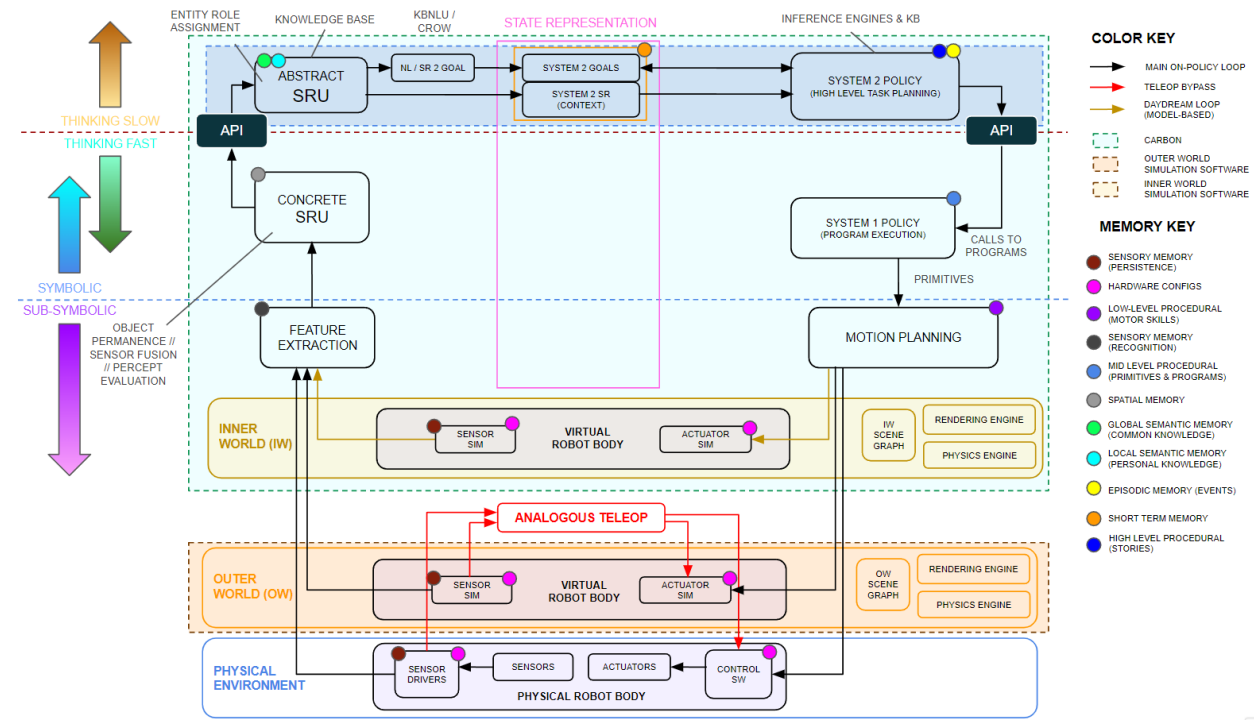

**Figure 23.** Added abstract state representation, full state representation, natural language to goal, and system 2 policy.

#### 4.5.2.11.1 Abstract State Representation Updater

We call the second part of the system's state representation the *abstract state representation*. The abstract state representation includes access to a large *knowledge base* providing a repository of common sense knowledge about the world, which in the current Carbon implementation is provided by the Cyc system (Lenat, 2023). This module is responsible for updating the abstract state representation using incoming concrete state representation information. This updating step also selects which parts of the knowledge base the system believes are relevant to the current situation.

#### 4.5.2.11.2 Natural Language to Goal State Representation

In the current Carbon implementation, *goal state representations* are logical statements, written in Cycorp's CycL language, that resolve to true if the system believes the goal state has been achieved.[4]

Goal state representations are defined over predicates that include perceptual information coming from the concrete state representation, including percept truth values and incoming natural language strings, and the abstract state representation, including common sense and semantic facts about the world inherited from the Cyc knowledge base and information extracted from memories. If the system doesn't have access to truth values of some of the predicates in the goal state representation, the goal state representation is assumed to be false.

Carbon's goal seeking behavior does not require natural language inputs, but it is influenced by them. As an example, the incoming natural language phrase "Did you touch a blue object located in the capital of France on September 25th, 2022?" is parsed to a CycL goal state representation which will resolve to true if the system responds verbally (and correctly) with either yes or no, depending on (a) whether a memory exists of touching a blue object; (b) knowledge of what the capital of France is (from the Cyc knowledge base), and whether this matches the location of the above event, which is obtained from memory; and (c) if both of the above are true, whether the event happened on that date, also pulled from memory. The first part of the statement (what the system responded verbally with, if anything) is accessed as a percept from the concrete state representation.

The CycL statement, which is the goal state representation for the issued incoming natural language phrase "Did you touch a blue object located in the capital of France on September 25th, 2022?", is explicitly:

```
Y = 
        (thereExists ?TALK
         (and
           (isa  ?TALK  Speaking)
           (doneBy  ?TALK  TheCrowSelf)
           (infoTransferred  ?TALK
              (thereExists ?OBJECT
                 (thereExists ?TOUCHING
                    (and
                       (dateOfEvent   ?TOUCHING   (CycLDateFn 2022 9 25))
                       (objectTouched   ?TOUCHING  ?object)
                       (isa  ?OBJECT  BlueColor)
                       (doneBy  ?TOUCHING  TheCrowSelf)
                       (eventOccursAt   ?TOUCHING  CityOfParisFrance)))))))
```

Y evaluates to true if the goal state has been met, and false otherwise.

---

[4] CycL is a logic-based language, designed to represent knowledge in a formal and structured way.

### 4.5.2.11.3   System 2 Policy

The system 2 policy module's job is to convert current state representation and goal state representation to task plans, which can be thought of as a traversal through state representation space from current to goal through intervening states via the serial application of Instruction calls. To do this, Carbon uses two pathways.

The first uses the Cyc system to both deduce and abduce, using logical inference, a task plan that takes the system from its current state to the goal state via sequential application of Instructions. This can be loosely conceptualized as being like a theorem prover, where the system attempts to prove a series of action-taking steps leads to the goal state representation being true. This includes action-taking steps that seek access to predicate truth values that are not available. For example, if the system is being asked if the bin on the table is blue, and it hasn't seen the bin, it may choose to look around first, or turn a particular feature extractor on, or load a particular vision model.

The second uses a Large Language Model (LLM) whose prompt is structured to contain relevant state representation information, goal state, and Instruction Set information, that generates task plans as sequences of Instructions. In the tests described in Section 5 the system used the OpenAI davinci-003 LLM. In this mode, Cyc is used to verify that each step produced by the LLM adheres to certain common sense constraints, including safety constraints, and to reason about whether the entirety of the proposed task plan if successfully executed would make the goal state true.

The abstract state representation contains a conditional flag which can be set to use one or both of these. For example, the system can use the first pathway if in a circumstance where careful reasoning is required (such as choosing moves in a chess game) or if we need high levels of safety and explainability (for example, using a dangerous machine), and the second if fast action-taking and low levels of safety are acceptable (for example, the sortation task described in the next section).

For the subtasks run in Section 5 in autonomous mode, we ran both in parallel, and the first successful task plan generated was used.

### 4.5.2.11.4   A Pass Through the Thinking Slow Pathway

Here is an example of a typical pass through the above machinery. Carbon has a state representation, which includes the concrete state representation. The concrete state representation contains all of the features extracted from the current situation, including any text received from speakers in the scene.

Carbon associates text with its origin (which could originate for example via OCR from a book, or via speech from a person). If the origin is speech from a person that Carbon recognizes, and that person is authorized to issue commands, then the text that person speaks is passed to the natural language to goal state representation module. Information about the person, including whether the system knows them and their authorization levels, is stored in the abstract state representation.

For this example, let's assume that the concrete state representation is consistent with the image in Figure 24. We recognize this as a robot situated in front of a chess board with some pieces on the board, with two blue bins situated to the back right and back left of the board. We then assume that an authorized person speaks a phrase; this phrase can be anything. This phrase is converted from audio to text in the feature extraction module. As a specific example, let's say the phrase is "Drop all the white pieces in one blue bin and all the black pieces in the other blue bin."

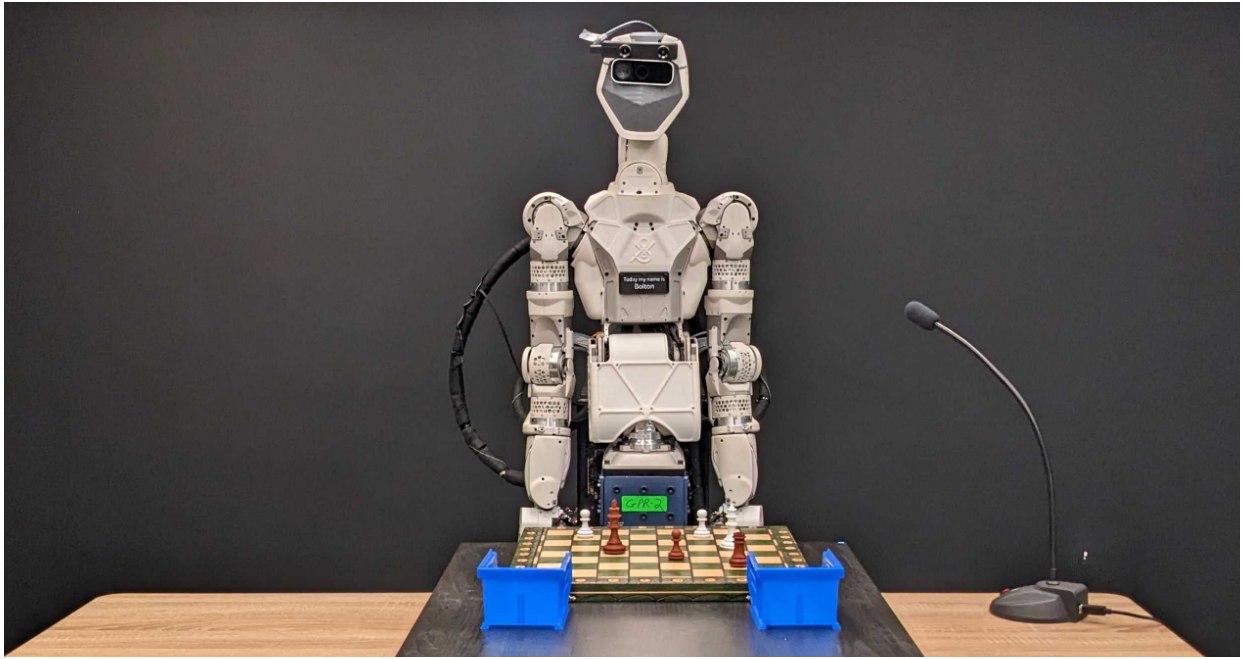

**Figure 24.** A robot situated in front of some chess pieces on a chess board with two blue bins.

The system 2 policy then takes as input the state representation, the goal state representation, and the available Instruction Set calls, and then attempts to generate task plans using both pathways available to the system 2 policy module.

Once a task plan has been completed and checked, it is pushed down through the HLC API to the system 1 policy system for step by step execution.

Here is an example of a successful task plan generated using the second pathway, using the natural language command and information from the state representation in a scenario similar to that shown in Figure 24 that the system inferred was relevant to the current situation.

When LLMs are involved, we use a process that first decimates the potential actions relevant for a task and then explicitly include the subset of Instructions believed to be sufficient in the LLM prompt. For this case, the final LLM prompt the system generated was:

> My name is Radcliffe. I am a humanoid general purpose robot. I am speaking with Geordie. I see the following objects: a white rook chess piece number 48, a white pawn chess piece number 80, a white pawn chess piece number 60, a black rook chess piece number 99, a black pawn chess piece number 8, a black pawn chess piece number 77, a blue bin number 13, a blue bin number 78. Geordie says 'Drop all the white pieces in one blue bin and all the black pieces in the other blue bin.' My goal is to do what Geordie asked me to do. I can only perform one action at a time. I will speak the phrase "task completed" when done.
> I can only choose actions from the following list:
> 1. look at object
> 2. grasp object
> 3. place object on object
> 4. speak a phrase
> Here is an enumerated list of the above actions I should perform in order to complete my goal:

The LLM returns a raw task plan, which is parsed to Instruction Set elements. This sequence of Instructions then undergoes an inspection process from the Cyc system that examines the logical consequences of each step in the context of the abstract state representation, and flags any steps that may require a person to check.

The raw task plan returned, the parsing to Instruction Set elements, and the result of the Cyc checks on each, are shown in Figure 25. In this particular task plan there were no flags raised; but unfortunately, many task plans returned by LLMs contain steps that violate basic common sense principles.

As an example of this, here are steps 4 and 8 of a task plan returned while executing one of the work subtasks described in Section 5, where the context is meal preparation using available objects.

> 4. drop laundry detergent 2 into pot 34
> 8. place pot 34 on table 1

While this was a valid task plan that could be executed, and does deliver a 'meal' to a person sat at the table, Cyc was able to reason that step 4 uses an ingredient that is not a valid part of a meal and flagged this for a person to attend to.

| Raw LLM output | Parsed to Instruction | Cyc Check |
|---|---|---|
| 1. look at white rook chess piece number 48 | set_self_look_at_object("48") | ✓ |
| 2. grasp white rook chess piece number 48 | set_self_grasp_object("48") | ✓ |
| 3. look at blue bin number 13 | set_self_look_at_object("13") | ✓ |
| 4. place white rook chess piece number 48 on blue bin number 13 | set_self_place_object_on("48", "13") | ✓ |
| 5. look at white pawn chess piece number 80 | set_self_look_at_object("80") | ✓ |
| 6. grasp white pawn chess piece number 80 | set_self_grasp_object("80") | ✓ |
| 7. look at blue bin number 13 | set_self_look_at_object("13") | ✓ |
| 8. place white pawn chess piece number 80 on blue bin number 13 | set_self_place_object_on("80", "13") | ✓ |
| 9. look at white pawn chess piece number 60 | set_self_look_at_object("60") | ✓ |
| 10. grasp white pawn chess piece number 60 | set_self_grasp_object("60") | ✓ |
| 11. look at blue bin number 13 | set_self_look_at_object("13") | ✓ |
| 12. place white pawn chess piece number 60 on blue bin number 13 | set_self_place_object_on("60", "13") | ✓ |
| 13. look at black rook chess piece number 99 | set_self_look_at_object("99") | ✓ |
| 14. grasp black rook chess piece number 99 | set_self_grasp_object("99") | ✓ |
| 15. look at blue bin number 78 | set_self_look_at_object("78") | ✓ |
| 16. place black rook chess piece number 99 on blue bin number 78 | set_self_place_object_on("99", "78") | ✓ |
| 17. look at black pawn chess piece number 8 | set_self_look_at_object("8") | ✓ |
| 18. grasp black pawn chess piece number 8 | set_self_grasp_object("8") | ✓ |
| 19. look at blue bin number 78 | set_self_look_at_object("78") | ✓ |
| 20. place black pawn chess piece number 8 on blue bin number 78 | set_self_place_object_on("8", "78") | ✓ |
| 21. look at black pawn chess piece number 77 | set_self_look_at_object("77") | ✓ |
| 22. grasp black pawn chess piece number 77 | set_self_grasp_object("77") | ✓ |
| 23. look at blue bin number 78 | set_self_look_at_object("78") | ✓ |
| 24. place black pawn chess piece number 77 on blue bin number 78 | set_self_place_object_on("77", "78") | ✓ |
| 25. speak the phrase "'task completed'" | set_self_say_text_phrase("'task completed'") | ✓ |
| Cyc check vs goal state representation | | ✓ |

**Figure 25.** Task plans are sequences of Instructions, analogous to a compiled program. Cyc checks the pre- and post-conditions of each Instruction to ensure that no violations of common sense or other constraints (such as safety) occur. In this case, Cyc affirms that each step is acceptable. In cases where steps fail, we have isolated the specific failure steps and can re-plan.

The combination of an LLM for open world task planning over the current state representation of the system, Cyc checking these task plans for violations of common sense, and a symbolic instruction set to compile down to that covers a large number of complex actions, allows the system to exhibit a large range of goal seeking behaviors.

Finally we add the features shown in Figure 26. This is the complete Carbon cognitive architecture.

### 4.5.2.12 Motion Goals, Joint States, & Planning Scene

The third part of the system's state representation includes a set of sub-symbolic features specifically necessary for motion planning. These include robot joint states, aspects of the scene required for the motion planning policies, including locations and orientations of objects, and target joint state trajectories.

### 4.5.2.13 System 1 Goals, Percepts, and Scene Graph

These three are all parts of the concrete state representation, and are split out in the diagram to show specifically how these different parts communicate with other parts of the architecture.

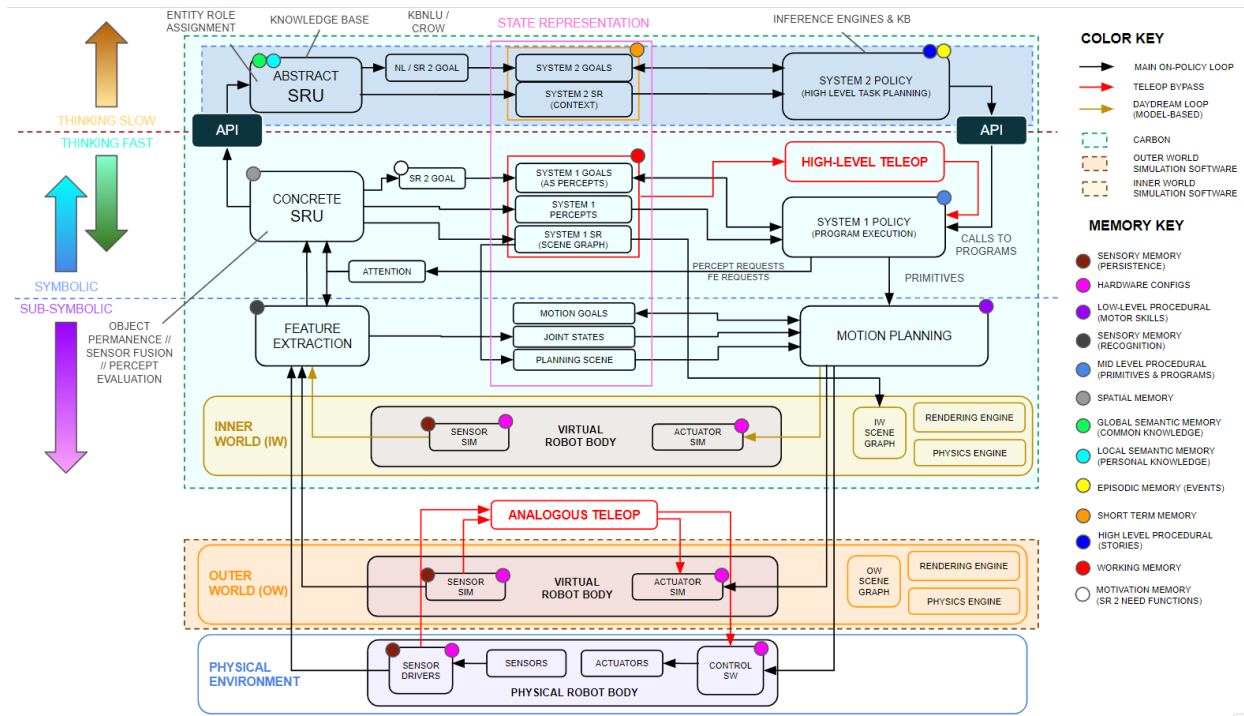

**Figure 26.** Added SR2Goal, system 1 goals, percepts, scene graph, attention, high level tele-op, working memory and motivation memory, paths to planning scene and IW scene graph.

### 4.5.2.14      Attention

The attention module turns feature detectors on and off, based on requests from the system 1 policy.

### 4.5.2.15      High Level Teleoperation

In analogous teleoperation, a person immersively and analogously controls a robot. This provides exceptional control over a robot. But it has a fundamental flaw from the perspective of transitioning to autonomous control. While the pilot and the autonomous system technically have access to the same sensory and actuation feeds, in practice this is not true. Looking carefully at the architecture in Figure 26, what the autonomous system really has access to on the action side are Instruction Set calls with symbolic parameters, and on the perception side the state representation and the inner world model.

High level teleoperation limits the interface of a person controlling a robot to being able to only select actions from the Instruction Set, and only being able to have access to state representation and inner world model information. Sensory information is displayed on a screen and actions are triggered by clicking icons on a screen. When actions are selected, the policy for accomplishing that action is triggered and the system autonomously attempts to complete the selected action.

A useful analogy is that high level teleoperation is like a point and click adventure game. In the analogy, the types of things you can click on are the Instruction Set elements and their parameters ("grasp(cup)") — the pilot can only interact with symbolic descriptions of the world. We do not allow pilots to see direct feeds from the outer world model (such as direct camera feeds) as these are not available to the Carbon architecture. Instead pilots can only see the inner world model of the robot — in other words, what the robot sees 'in its mind'. This allows us to understand better why the autonomous system may be failing. For example, it just may not be creating a good enough inner world model.

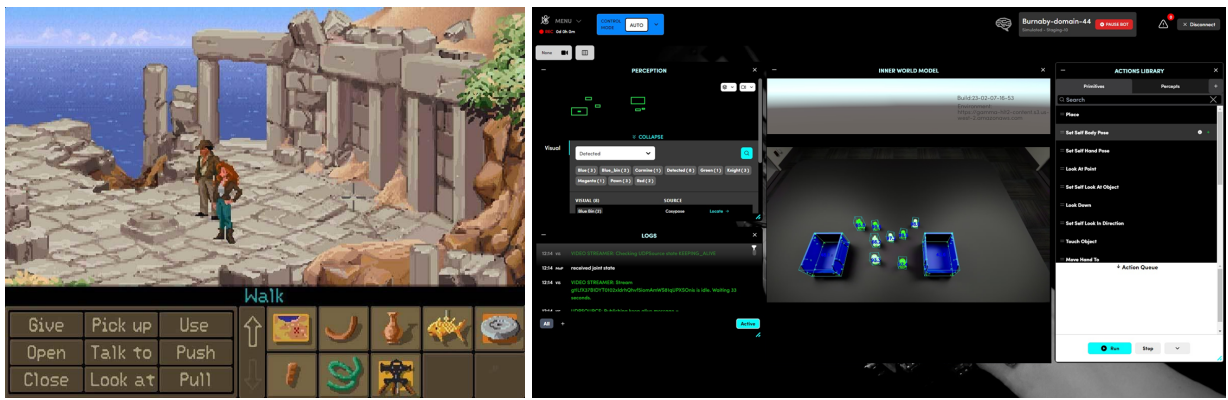

**Figure 27**. Left: An example point and click game interface (from the game "Indiana Jones and the Fate of Atlantis" by LucasArts). The buttons bottom left are analogous to Instructions and the objects bottom right are analogous to the parameters in those Instructions. Right: a screenshot of the Sanctuary high level teleoperation system. The inner world model is visualized center middle, symbolic features extracted are shown on the left, and Instructions are on the right.

A practical advantage of high level teleoperation is that it is performed using a keyboard, screen, and mouse. It does not require any specialized gear. As it is by its nature run remotely through a browser, this type of control can be scaled arbitrarily. Anyone anywhere in the world can control a robot using this style of control, whereas analogous teleoperation is constrained to lab conditions.

### 4.5.2.16    Memory Types and Definitions

In the study of the brain, many different kinds of memories have been identified [(Atkinson, 1968)](#). In Carbon, 13 memory types have been implemented. Some of these are roughly analogous to human memory types and when that is the case we use similar naming conventions. While Carbon is running, all memories, regardless of type, are labeled and stored for potential downstream and offline use.

1. **Sensory memory** in humans is a very short-term memory store for information being processed by the sense organs. In people, sensory memory has a limited duration to store information, typically less than a second. In Carbon, there are two types of sensory memory implemented. The first (labeled *persistence* in the key) stores all incoming raw sensor data (video, audio, proprioceptive and haptic). The second (labeled *recognition* in the key) stores all features extracted by the feature extraction module.

2. **Hardware Configs**, or equivalently **Hardware Abstraction Layer**, or **HAL**, is a type of memory that stores facts about the physical nature of the specific robot being controlled, such as servo names and types, camera properties, and the like. It is a type of semantic memory, but for technical reasons we keep it distinct from other types of semantic memory.

3. **Procedural memory** is a type of implicit memory (unconscious, long-term memory) which aids the performance of particular types of tasks without conscious awareness of these previous experiences. Procedural memory guides the processes we perform, and most frequently resides below the level of conscious awareness. When needed, procedural memories are automatically retrieved and utilized for execution of the integrated procedures involved in both cognitive and motor skills, from tying shoes, to reading, to flying an airplane. Procedural memories are accessed and used without the need for conscious control or attention. Procedural memory is created through procedural learning, or repeating a complex activity over and over again until all of the relevant neural systems work together to automatically produce the activity. Implicit procedural learning is essential for the development of any motor skill or cognitive activity. In Carbon, we distinguish between three different types of procedural memory. High level procedural memory stores task plans generated by the system 2 policy, which include natural language descriptions of goals and task plans; medium level procedural stores sequences of Instructions executed by the system 1 policy; and the low level procedural memory stores sequences of joint states executed by the motion planning module.

4. **Spatial memory** refers to memory of locations and properties of a scene. In Carbon, we store sequences of concrete state representations in this memory type.

5. **Semantic memory** refers to a portion of long-term memory that processes ideas and concepts that are not drawn from personal experience. Semantic memory includes things that are common knowledge, such as the names of colors, the sounds of letters, the capitals of countries and other basic facts acquired over a lifetime. In Carbon, we implement both local semantic memory, which is stored sequences of abstract state representations, and global semantic memory, which is a repository of knowledge about the world stored in a separate repository (in the current implementation, this is the Cyc knowledge base).

6. **Episodic memory** is a category of long-term memory that involves the recollection of specific events, situations, and experiences. Your memories of your first day of school, your first kiss, attending a friend's birthday party, and your brother's graduation are all examples of episodic memories. In Carbon, episodic memories are stored sequences of task plans that were actually executed.

7. **Short-term memory** (or "primary" or "active memory") is the capacity for holding, but not manipulating, a small amount of information in mind in an active, readily available state for a short period of time. For example, short-term memory can be used to remember a phone number that has just been recited. The duration of short-term memory (when rehearsal or active maintenance is prevented) is believed to be in the order of seconds. In Carbon, short-term memory is used as a scratch pad for holding information relevant to system 2 processes.

8. **Working memory** is a cognitive system with a limited capacity that can hold information temporarily. Working memory is important for reasoning and the guidance of decision-making and behavior. In Carbon, working memory is used as a scratch pad for holding information relevant to system 1 processes.

**4.6    Implementing Evolutionary Learning With The Workflow Process**

Carbon as described in Section 4.5 lacks explicit learning mechanisms. It satisfies the first part of Nilsson's habile system requirement, but not the second part. Here we address this by introducing an evolutionary learning system which we call the *workflow process*.

The workflow process (shown in Figure 28) is designed to evolve a robot and its control system towards being able to autonomously complete a growing number of work subtasks. It identifies what part of the system is preventing the system from succeeding, flagging these for upgrades by teams of human engineers.

Here we describe each step of the workflow process. The workflow process requires that all preceding steps fully pass before the next step can begin. This means that getting a subtask through the entire process generally requires multiple backtracking steps as blocking problems arise that require system upgrading / evolution.

**4.6.1    Step 1: Data Capture From People Doing Work**

The first step's goal is to fully characterize how work is done by people. We implemented a system where people, who we refer to as Field Service Associates (FSAs), are given verbal instructions, and then perform requested work tasks while wearing cameras recording their first person visual experience. In addition to the camera feeds, FSAs create structured text-based accounts of work activities performed together with the verbal instructions given, and all task timing is recorded.

The raw data collection from FSA work activities is Step 1A of the workflow process. Figure 29 shows an FSA performing work tasks (note the camera mounted on her head).

If some aspect of the raw data collection creates downstream problems, this is reported to the FSA Operations team, which is responsible for managing all aspects of the FSA work.

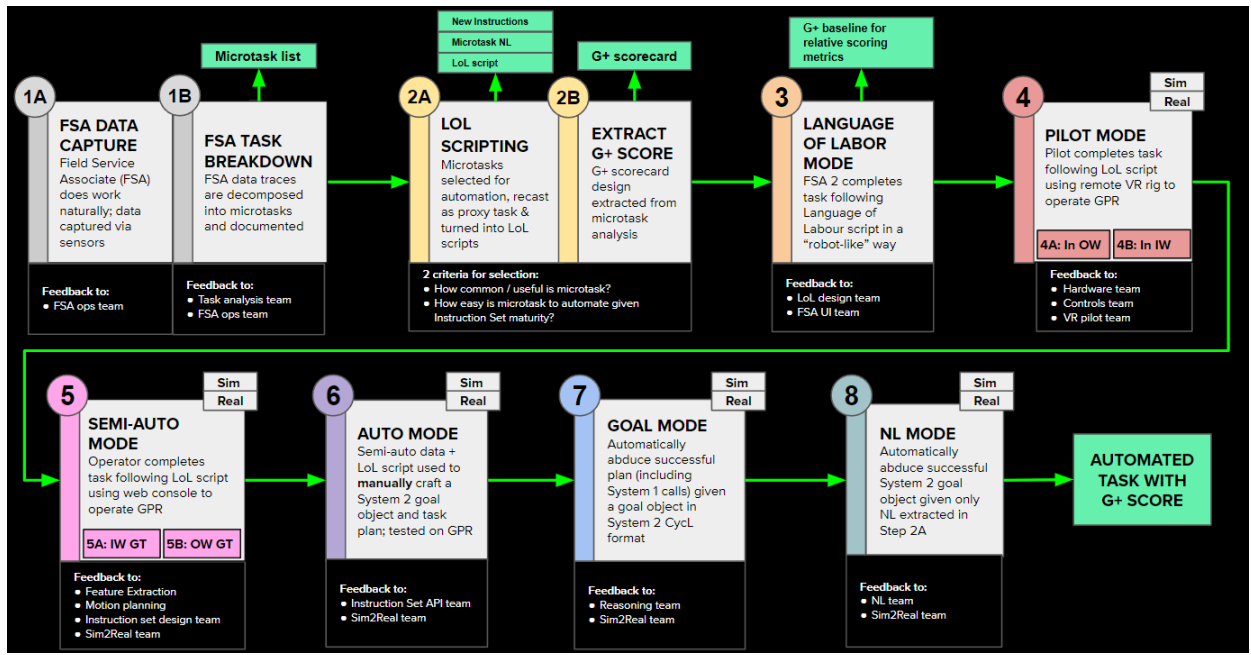

**Figure 28.** The workflow process is a systematic evolutionary learning process acting on all parts of the humanoid robot and Carbon system.

The data received from this work is then processed by a task analytics team, which is responsible for identifying subtasks within the work done that would make good downstream subtask automation targets. What makes a subtask a 'good automation target' is largely down to two things: (a) do we believe the current system can, currently or with minor modification, automate it, by which we mean generate and successfully execute a sequence of Instructions that achieves the goal of the subtask; and (b) how much of the full work cycle consists of repeating blocks of that subtask. For example, in the scenario in Figure 29, verbally interacting with a customer is repeated on the order of 100 times per eight hour shift, and it can be done using sequences of currently implemented Instructions, making it a good subtask.

The identification and prioritization of subtasks is Step 1B of the workflow process. If these subtasks cause downstream issues, the task analytics team is responsible for fixing these issues. The output of Step 1B is a subtask list.

### 4.6.2   Step 2: Subtask Breakdown Into Task Plans Written as Sequences of Instruction Set Calls

Given a subtask from Step 1, Step 2 then uses people to decompose the subtask into a series of action-taking steps, each of which is an allowed Instruction drawn from the current Instruction Set. This is a 'human compiler' step. If the people working to do this cannot create a sensible task plan to achieve the subtask, this indicates that the Instruction Set lacks some key element required. This feedback is sent to the team responsible for the Instruction Set, where a new Instruction Set element is engineered (or learned), unblocking Step 2.

If a task plan can be created to potentially perform the subtask using only allowed Instructions, it is returned as the output of Step 2A.

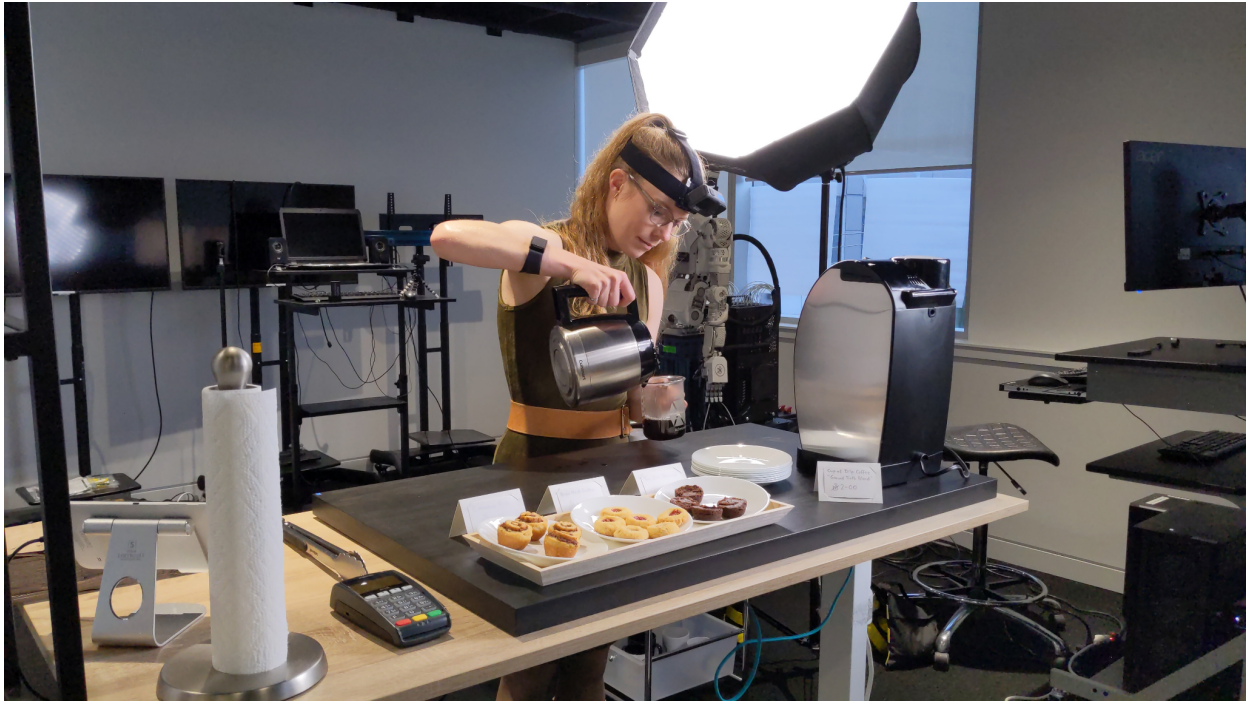

**Figure 29.** Step 1A of the workflow process collects first person data from FSAs performing work.

Step 2B takes the subtask in question, and creates a subtask fingerprint and associated g+ score, using the process described in Section 2. This step never gets blocked, so there is no failure feedback mechanism for this.

### 4.6.3   Step 3: FSA Attempts to Execute Task Plan Created in Step 2

In Step 3 we verbally repeat the task plan generated in Step 2 back to an FSA who has not encountered the subtask before. This step tests whether a person, following the task plan written as a sequence of Instructions, can perform the subtask. In this mode, the FSA being asked to perform each Instruction may not have enough information to perform a particular step, or the step may be parametrized incorrectly, or there may be other failure modes. FSAs at this step are asked to be 'robot-like humans' and follow each Instruction as literally as possible, as if they were the robot.

If the FSA is unable to complete the full subtask task plan satisfactorily, the reasons for this are communicated to the team generating task plans in Step 2, who then attempt to change the task plan to adjust to the feedback. This may entail altering the task plan using the existing Instruction Set, or may require new Instructions to be created.

During this step, subtask performance characterization is done, which includes timing and work quality measurements.

### 4.6.4 Step 4: Analogous Teleoperation Mode

In Step 4 we repeat the task plan generated in Step 2 back to a pilot, who is operating a robot using analogous teleoperation. This step tests whether a person can complete the task plan using a robot under analogous teleoperation. What is being tested here is whether the robot is physically capable of executing each of the steps in the task plan.

Step 4 is done in four different modes. Step 4A performs the step in both simulation and in reality, where in simulation the pilot has access to feeds from the outer world model. Step 4B performs the step in both simulation and in reality, but in this case the pilot only has access to feeds from the robot's inner world model. Step 4B additionally tests whether the robot's inner world model is sufficient to allow a human pilot to perform the task; this step can cause failures if, for example, the systems responsible for populating objects into the inner world are placing or orienting them incorrectly.

If any step of the task plan cannot be performed by the simulated robot under analogous teleoperation, this means that one or more of the simulation, controls, or teleoperation systems are insufficient and require upgrading. If any step of the task plan cannot be performed by the physical robot under analogous teleoperation, this means that one or more of the hardware, controls, or teleoperations systems are insufficient and require upgrading. For each failure case, feedback is provided to the simulation, hardware, controls, or teleoperations teams, who are responsible for upgrading the systems in question that block success.

During this step, subtask performance characterization is done in all control modes, which includes timing and work quality measurements. The expectation is that timing and work quality will both degrade from step 3 to step 4, as immersively controlling a robot body is considerably more difficult than controlling your own body.

### 4.6.5 Step 5: High Level Teleoperation Mode

In Step 5 we repeat the task plan generated in Step 2 back to a pilot, who is operating a robot using the high level teleoperation interface, which exposes Instruction Set commands as clickable graphical elements on a screen. This step tests whether the autonomous policies for completing each Instruction, triggered in series by a person through the high level teleoperation interface, are successfully executing each of the steps in the task plan.

Step 5 is done in the same four modes as Step 4. In this case, the main failure modes tested are the autonomous policies for completing each Instruction. In the case where one or more of these fail, this information is sent to one or more of the controls, feature extraction, motion planning, or policy teams, depending on the specific failure.

As in the previous steps, subtask performance characterization is done in all control modes.

### 4.6.6 Step 6: Autonomous Mode

In Step 6 we run the task plan generated in Step 2 as a fully autonomous script, where the system executes the entire task plan as a sequence of autonomous policies each implementing one Instruction call. This is done in both simulation and physical reality. If this step successfully executes in both, we consider the subtask automated, with the caveats that (a) it may not be clear to what extent the success generalizes to other similar situations, and (b) the task plan was written by a person.

As in the previous steps, subtask performance characterization is done in all control modes.

### 4.6.7 Step 7: Goal Mode

In Step 7 we provide a text string to the system, which is intended to be a command to achieve a goal, and a hand-crafted goal state representation written as a Boolean CycL statement. Step 7 tests whether the system 2 task planner can autonomously generate a task plan that can be successfully executed in both simulation and reality to achieve the goal state, in essence short-circuiting most of the previous steps, going from a verbal instruction directly to an executable task plan. This primarily tests the first pathway of the system 2 policy — Cyc's ability to logically deduce and/or abduce a task plan given an input command and the abstract state representation of the system.

As in the previous steps, subtask performance characterization is done in all control modes.

### 4.6.8 Step 8: NL Mode

In Step 8 we provide a text string to the system, which is intended to be a command to achieve a goal, but not the hand-crafted goal state representation, which must be automatically generated by the natural language to goal state representation module. Step 8 additionally tests whether the full system 2 policy, comprising both pathways, can generate task plans that can be successfully executed in both simulation and reality.

As in the previous steps, subtask performance characterization is done in all control modes.

# 5. Extracting Work Fingerprints and g+ From Subtask Completion

O*NET provides occupation fingerprints, which are empirically extracted using the process described in Section 2 (practitioners are asked what levels of each of the fingerprint dimensions are minimally required to perform the occupation). These are available in the *O*NET Occupations Ranked by g+* tab in (Data, 2023). The average g+ over occupations is 100 by definition, and has a standard deviation of 15.4. The occupation with the lowest g+ is Models (44.7), whereas the highest is First-Line Supervisors of Firefighting and Prevention Workers (141.3).

O*NET does not provide fingerprints for either detailed work activities or tasks. We extracted upper bounds for these using the following procedure. We first extract detailed work activity fingerprints by, for each detailed work activity, listing the occupations that O*NET lists as requiring it, and then computing the minimum values across fingerprint dimensions for those occupations. The reasoning is that the detailed work activity in question could not require more of any of the fingerprint dimensions than any of the occupations it requires. The estimated upper bounds for fingerprints for detailed work activities are listed in the *O*NET Detailed Work Activities Ranked by g+* tab in (Data, 2023). The average g+ over detailed work activities is 72.2, and has a standard deviation of 15.9. The detailed work activity with the lowest g+ is Clean Facilities or Work Areas (31.5), whereas the highest is Protect Property from Fire or Water Damage (135.9).

We then performed a similar analysis for tasks, where for each task we list the detailed work activities associated with it, and find the minimum values of all fingerprint dimensions across that set. This provides an estimated upper bound for each task. O*NET tasks are listed in the *O*NET Tasks Ranked by g+* tab in (Data, 2023). The average g+ over tasks is 64.4, and has a standard deviation of 14.8. The task with the lowest g+ is Perform Serving, Cleaning, Or Stocking Duties In Establishments, Such As Cafeterias Or Dining Rooms, To Facilitate Customer Service (30.2), whereas the highest is Take Action To Contain Any Hazardous Chemicals That Could Catch Fire, Leak, Or Spill (135.9).

The current portfolio of subtasks that have passed at least step four (successful completion under analogous teleoperation) in the workflow process are listed in the *Subtask Portfolio & Fingerprints* tab in (Data, 2023). This portfolio comprises 350 subtasks. Each comes with a Type and a Date. The Type labels different subsets of the full portfolio. The TRL-1, EOY-22 and LLM-1 subsets were all run up to step 6 (autonomous mode) in the workflow process, whereas the UAT subset were all run up to step 4 (analogous teleoperation). The Date field shows the date of the first time the system successfully accomplished that subtask. The average g+ values of these subtasks is 5.7, far below the g+ values of O*NET tasks.

We used Algorithm 1 to extract g+ values over time from this portfolio, for both analogous teleoperation and autonomous control modes. These are plotted in Figure 29. The current g+ value of the system under analogous teleoperation is 78.2, and under autonomous control 73.7. The extracted work fingerprints for the current system under both analogous teleoperation and autonomous control are shown in Figure 30.

Having extracted work fingerprints for the current system under different control modes, we then compare these to the 19,265 O*NET task fingerprints to see which, if any, we can infer the current system could perform. Our process infers that seven of the O*NET tasks are performable by the current system under analogous teleoperation, whereas none can be performed as yet under autonomous control. The O*NET tasks that are currently inferred to be performable under analogous teleoperation are shown in Figure 31.

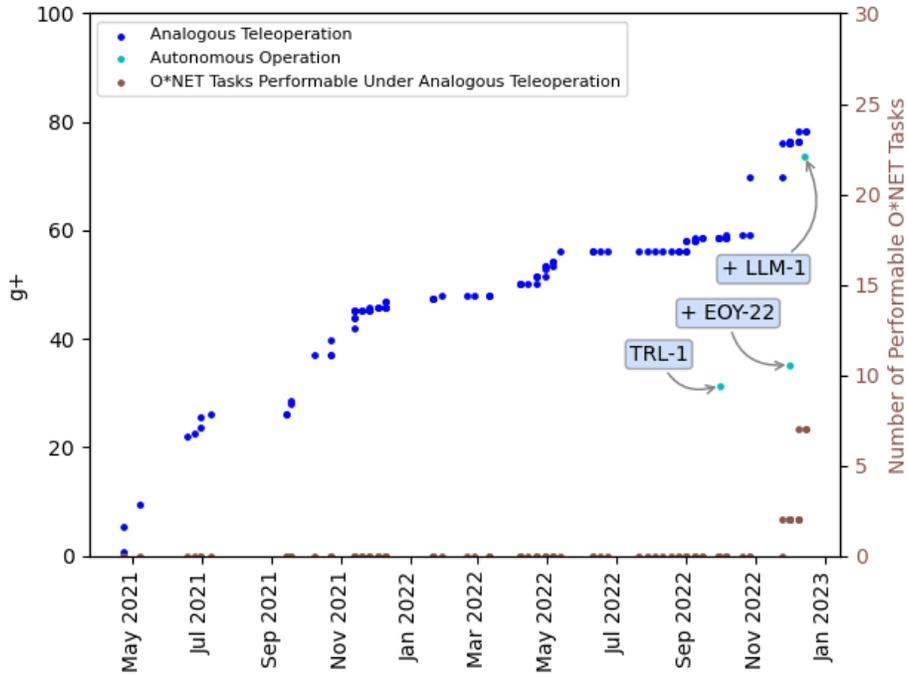

**Figure 29.** Historical g+ scores over time from running the workflow process under both analogous teleoperation (blue) and autonomous (cyan) control. The annotations on the autonomous control points indicate which subtask portfolios were successfully completed autonomously (see **Types** annotation in the *Subtask Portfolio & Fingerprints* tab (Data, 2023)). The g+ score of the human pilot who generated the blue points was 158.8. As the physical capabilities of the robot increase over time we would expect g+ under analogous teleoperation to eventually saturate near the pilot's g+ score. The brown points indicate how many O*NET tasks are performable under analogous teleoperation given the robot's work fingerprint over time. Currently no O*NET tasks are performable autonomously.

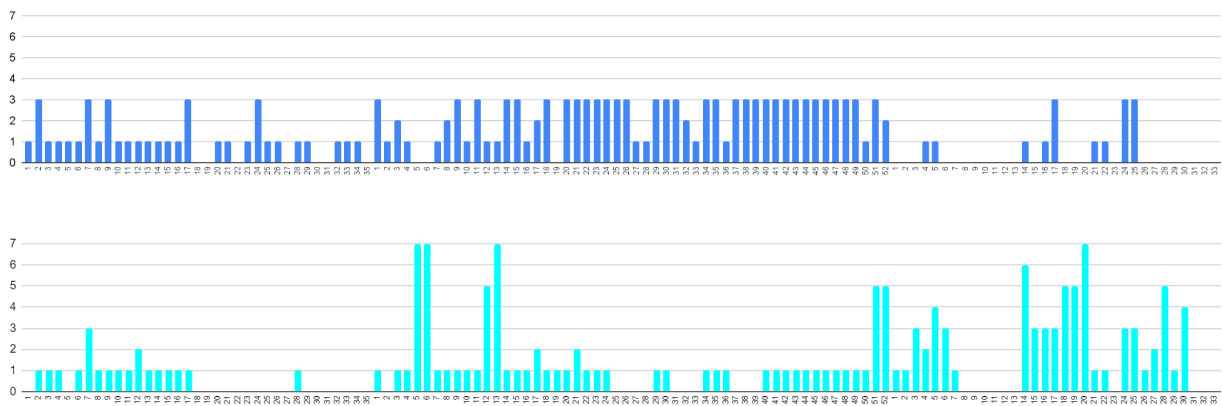

**Figure 30.** Extracted current work fingerprints for (top, blue) analogous teleoperation (step 4) and (bottom, cyan) autonomous control (step 6). The autonomous control system is superior to our human pilot in several cognitive abilities (primarily related to memory and mathematical ability) and knowledge categories (due to the knowledge contained in the Cyc and LLM systems used).

| Task ID | Occupation | Task |
|---|---|---|
| 657 | First-Line Supervisors of Retail Sales Workers | Provide customer service by greeting and assisting customers and responding to customer inquiries and complaints |
| 682 | Counter and Rental Clerks | Greet customers and discuss the type, quality, and quantity of merchandise sought for rental |
| 694 | Retail Salespersons | Greet customers and ascertain what each customer wants or needs |
| 779 | Medical Secretaries and Administrative Assistants | Greet visitors, ascertain purpose of visit, and direct them to appropriate staff |
| 2385 | Recreation Workers | Greet new arrivals to activities, introducing them to other participants, explaining facility rules, and encouraging participation |
| 2791 | Secretaries and Administrative Assistants, Except Legal, Medical, and Executive | Greet visitors or callers and handle their inquiries or direct them to the appropriate persons according to their needs |
| 20431 | Switchboard Operators, Including Answering Service | Greet visitors, log them in and out of the facility, assign them security badges, and contact employee escorts |

**Figure 31.** Currently seven (out of 19,265) O*NET tasks are performable under analogous teleoperation with the current system. None of the O*NET tasks are currently performable under autonomous control.

The g+ values of the current system are quite high compared to the required g+ values for O*NET tasks. However only a small number of these are performable. This is due to the unusual distribution of fingerprint dimensions in the current system (Figure 30). Some of them are quite high (generally those associated with cognitive processes, where the system inherits the cognition of the pilot or the numeracy or knowledge of the autonomous system), but many of the more physical dimensions are quite low or zero. Currently the system's capability for work is somewhat analogous to a physically disabled person with very good cognitive skills. This suggests that as the physical capabilities of the underlying system mature past certain thresholds, there will be a highly nonlinear increase in performable tasks. Note that the type of analysis we performed here, with only slight improvements to certain of the fingerprint dimensions, predicts thousands of performable tasks covering a significant portion of the economy.

The observed trend is an increase in g+ of roughly 4 per month over the past 20 months. If the trend in g+ increase can be maintained, this would lead to saturation of g+ around its maximum value of 314 in approximately 60 months. This is a rough lower bound on when we might expect to be able to perform all O*NET tasks.

# 6. Discussion and Limitations

In Section 2 we presented a framework for measuring progress towards human-level AI. In Section 5 we showed results of running that process on a system comprising a humanoid robot and a cognitive architecture-based control system. We described at a high level a current snapshot of the humanoid robot, the cognitive architecture used, and certain infrastructural pieces that support both. We also described a novel evolutionary learning algorithm, which we call the workflow process, which evolves this system under pressure to increase our measure of human-like intelligence.

We found that the g+ score of the current system under analogous teleoperation is approaching average human levels. This means that given fairly straightforward upgrades to the mobility, sensors, and hand dexterity of the system, it should be possible to perform a significant fraction of the world's work under analogous teleoperation.

The g+ of the system under full autonomous control is approximately the same, but the distribution of the work fingerprint in that case is significantly lagging in many important dimensions. What is required there is significant advancement in the stability and breadth of the Instruction Set, better and faster motion planners, and better and faster task planners.

This collection of technologies and ideas is very complex, and the approaches described here come with a host of issues and limitations. We conclude by listing some of the most important of these.

## 6.1 Work and Subtask Fingerprints and Measuring Progress Towards Human-Level AI

We have argued that explicitly defining the set of goals people seek is necessary for defining what human-like intelligence means. We use the set of goals presented to us when we do work as being a proxy for the set of all human-like goals. We then define human-like intelligence to be the capacity to achieve those goals, and define a measure of human-like intelligence (g+) for measuring progress on a scale that can be compared directly to human intelligence.

This approach has the following issues:

1. The choice of 120 latent variables selected by O*NET assumes many capabilities that people all natively have, but machines don't. This set could be augmented to include more basic latent variables that the current set takes for granted, such as fidelity of inner world models. This augmentation could make it easier to generate gradients towards human-level AI, and possibly reduce the impact of the following two issues.

2. The need for human practitioners to assign subtask fingerprints to subtasks. This is Step 2 in Algorithm 1, and a core part of this method. That step introduces subjectivity and lack of consistency into the process.

3. What it means to successfully accomplish a subtask is not always clear. People are very good at generalizing task performance between different instances of the same subtask, but this is not necessarily the case for machines. When a machine succeeds at performing a subtask, we are assuming here that the machine can do any instance of that subtask as well. But this is a dangerous assumption, and opens up possibilities for 'gaming' the test by creating systems that are exceptional at specific instances of subtasks, running the test only on those instances, and then claiming victory, when the spirit of the test assumes natural common sense generalization of the subtask. The underlying reason for the tests proposed here is to have a way to measure a progress gradient, not to score highly for the sake of just getting a high score. We have found that concretizing the subtask statements down to what the system's current generalization capabilities are known to be able to support has been a good strategy for avoiding this pitfall. For example, if the system is capable of grasping and moving a certain kind of object, and the Instructions involved in success are not functions of the color of the object, then we can safely word the subtask to include performing the task over different colored objects of the sort that are known to be graspable, but not extending the subtask definition to, for example, all objects.

## 6.2 The Physical Robot

Our expectation is that under analogous teleoperation, a humanoid robot should be able to match the g+ score of its pilot. While this seems reasonable, it has a rather remarkable implication — that such a system would be capable of performing a large fraction of the world's work.

In the case of the tests reported on here, the pilot had a g+ score of 158.8, while the measured g+ of the piloted robot was approximately 78. We have done a preliminary analysis of gaps between pilot and robot work fingerprints and have identified a set of improvements that will close the gaps. These include, in descending order of importance, improved dexterity and sensors on the hands; improved haptic and force feedback to the pilot; improved payload; improved visual feedback to the pilot; and moving from a wheeled and tethered base to untethered bipedal locomotion.

Our view is that while each of these presents challenges, enough has been demonstrated by the community in each of these categories to make closing the gap enough to unlock a large amount of work under analogous teleoperation straightforward.

## 6.3 The Cognitive Architecture

Our strategy of defining and engineering an Instruction Set, and then achieving goals via task plans constructed of sequences of Instructions, rests on three key capabilities. The first is building a robust and broad Instruction Set. The second is being able to identify aspects of state representation with arbitrary descriptors of the world, including goals, stated in natural language. The third is to create systems that can create task plans formed of sequences of Instructions to achieve the system's goals.

The cognitive architecture described in Section 4 is our current best implementation of a system that delivers those three capabilities, and all of the infrastructure required to support them. Many aspects of the Carbon cognitive architecture can be significantly improved. Here are some of the limitations we currently face:

1. Our pipeline for building object-centric world models requires the creation, by people, of models of the objects in the robot's environment that the robot can then detect, localize, and act on (for example by grasping them). This is clearly not scalable. There is a requirement to be able to build perception systems that are able to, on the fly and in general, populate objects into an inner world without having seen those objects before. While there are aspects in the current architecture (such as SLAM algorithms) that have this character, eventually all aspects of the visual system must adopt an 'on the fly' paradigm.

2. Our current method of linking state representation to verbal descriptions of the world is brittle and requires further development. Somewhat analogously to the problem identified above, currently mapping natural language statements about goals to concrete state representations have people

      in the loop in many parts of the process. Ultimately we would like to have a principled fully automated way to do this.

3. Improving the task planning process, given a fixed Instruction Set, is required. The problem is somewhat analogous to building both automated programming (program construction from goal specification) and compilation (to Instructions) tools.

4. Physics engines in both inner and outer worlds must be significantly improved. Specifically, the physical simulation of softbody contact physics in and around the hands is not good enough. This impacts simulation of both many parts of the world and, critically, the interaction of the robot's hands with the world. The hands of the robots have deformable parts that aid in grasping. Being able to properly simulate the contact physics of the hands with their environment is required for high fidelity simulation and sim2real transfer.

## Acknowledgements

The authors would like to thank Rich Sutton and Doug Lenat for their comments on this manuscript.

## O*NET